\documentclass[sigconf]{acmart}

\usepackage{booktabs} 
\usepackage{amsmath,amsfonts,epsfig}
\usepackage{textcomp}
\usepackage{verbatim}
\usepackage{url}
\usepackage{tablefootnote}

\usepackage{enumitem}
\usepackage{balance}

\newlist{myitemize}{enumerate}{10}
\setlist[myitemize]{label*=(\arabic*),nosep,leftmargin=*}

\copyrightyear{2017}
\acmYear{2017}
\setcopyright{acmcopyright} 
\acmConference{MM '17}{}{October 23--27, 2017, Mountain View, CA, USA.} 
\acmPrice{15.00} 
\acmDOI{https://doi.org/10.1145/3123266.3123359}
\acmISBN{ISBN 978-1-4503-4906-2/17/10}

\begin{document}
	\fancyhead{}
	\settopmatter{printacmref=false, printfolios=false}
\title{NormFace: $L_2$ Hypersphere Embedding for Face Verification}

\author{Feng Wang}
\authornote{Alan L. Yuille's visiting student.}
\affiliation{%
  \institution{University of Electronic Science and Technology of China}
  \streetaddress{2006 Xiyuan Ave.}
  \city{Chengdu} 
  \state{Sichuan} 
  \postcode{611731}
}
\email{feng.wff@gmail.com}

\author{Xiang Xiang}
\affiliation{%
\institution{Johns Hopkins University}
\streetaddress{3400 N. Charles St.}
\city{Baltimore} 
\state{Maryland} 
\postcode{21218}
}
\email{xxiang@cs.jhu.edu }

\author{Jian Cheng}
\affiliation{%
	\institution{University of Electronic Science and Technology of China}
	\streetaddress{2006 Xiyuan Ave.}
	\city{Chengdu} 
	\state{Sichuan} 
	\postcode{611731}
}
\email{chengjian@uestc.edu.cn}

\author{Alan L. Yuille}
\affiliation{%
  \institution{Johns Hopkins University}
  \streetaddress{3400 N. Charles St.}
  \city{Baltimore} 
  \state{Maryland} 
  \postcode{21218}
}
\email{alan.yuille@jhu.edu }

\renewcommand{\shortauthors}{Feng Wang, Xiang Xiang, Jian Cheng, Alan L. Yuille}

\begin{abstract}
Thanks to the recent developments of Convolutional Neural Networks, the performance of face verification methods has increased rapidly. In a typical face verification method, feature normalization is a critical step for boosting  performance. This motivates us to introduce and study the effect of normalization during training. But we find this is non-trivial, despite normalization being differentiable. We identify and study four issues related to normalization through mathematical analysis, which yields understanding and helps with parameter settings. Based on this analysis we propose two strategies for training using normalized features. The first is a modification of softmax loss, which optimizes cosine similarity instead of inner-product. The second is a reformulation of metric learning by introducing an agent vector for each class. We show that both strategies, and small variants, consistently improve performance by between 0.2\% to 0.4\% on the LFW dataset based on two models. This is significant because the performance of the two models on LFW dataset is close to saturation at over 98\%.
\end{abstract}

%
%




\maketitle
\section{Introduction}
%
%

In recent years, Convolutional neural networks (CNNs) achieve state-of-the-art performance for various computer vision tasks, such as object recognition \cite{krizhevsky2012imagenet, simonyan2014very, szegedy2014going}, detection \cite{girshick2014rich}, segmentation \cite{LongSD15Fully} and so on. In the field of face verification, CNNs have already surpassed humans' abilities on several benchmarks\cite{taigman2014deepface, lu2014surpassing}.

The most common pipeline for a face verification application involves face detection, facial landmark detection, face alignment, feature extraction, and finally feature comparison. In the feature comparison step, the cosine similarity or equivalently $L_2$ normalized Euclidean distance is used to measure the similarities between features. The cosine similarity $\frac{\langle \cdot , \cdot \rangle}{\| \cdot \|\|\cdot \|}$ is a similarity measure which is independent of magnitude. It can be seen as the normalized version of inner-product of two vectors. But in practice the inner product without normalization is the most widely-used similarity measure when training a CNN classification models \cite{krizhevsky2012imagenet, simonyan2014very, szegedy2014going}. In other words, the similarity or distance metric used during training is different from that used in the testing phase. To our knowledge, no researcher in the face verification community has clearly explained why the features should be normalized to calculate the similarity in the testing phase. Feature normalization is treated only as a trick to promote the performance during testing.

To illustrate this, we performed an experiment which compared the face features without normalization, \emph{i.e.} using the unnormalized inner-product or Euclidean distance as the similarity measurement. The features were extracted from an online available model \cite{wen2016discriminative}\footnote{\url{https://github.com/ydwen/caffe-face}}. We followed the standard protocol of \emph{unrestricted with labeled outside data}\cite{huang2014labeled} and test the model on the Labeled Faces in the Wild (LFW) dataset\cite{huang2007labeled}. The results are listed in Table \ref{tab:feature_norm}.

\begin{table}[!htb]
	\caption{Effect of Feature Normalization}
	\label{tab:feature_norm}
	\begin{tabular}{ccc}
		\toprule
		Similarity &Before Normalization&After Normalization\\
		\midrule
		Inner-Product & 98.27\% & 98.98\%\\
		Euclidean & 98.35\% & 98.95\%\\
		\bottomrule
	\end{tabular}
\end{table}

As shown in the table, feature normalization promoted the performance by about 0.6\% $\sim$ 0.7\%, which is a significant improvement since the accuracies are already above 98\%. Feature normalization seems to be a crucial step to get good performance during testing.  Noting that the normalization operation is differentiable, there is no reason that stops us importing this operation into the CNN model to perform end-to-end training.

\begin{figure}
	\centering
	\includegraphics[scale=0.3]{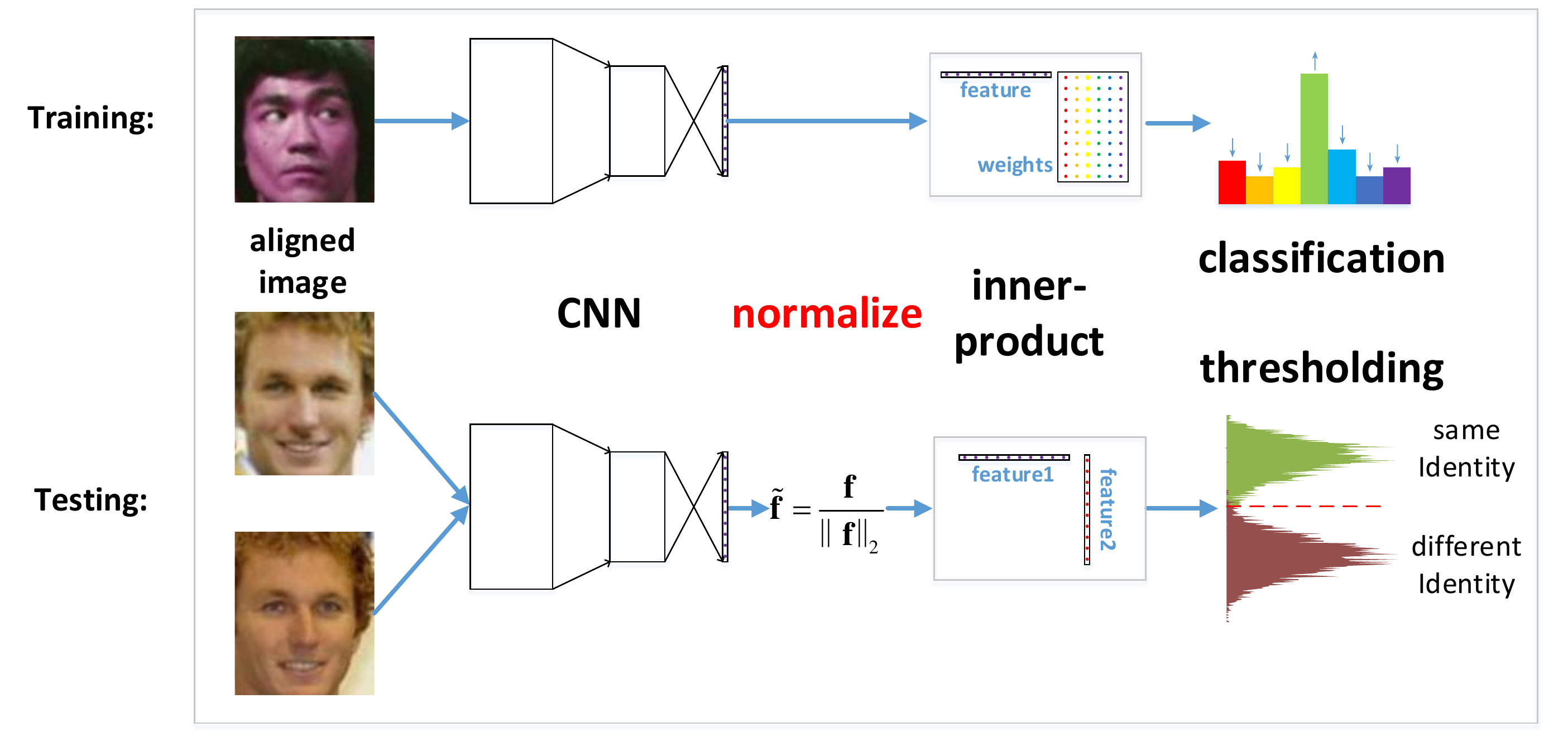}
	\caption{ Pipeline of face verification model training and testing using a classification loss function. Previous works did not use the normalization after feature extraction during training. But in the testing phase, all methods used a normalized similarity, e.g. cosine, to compare two features. }
	\label{fig:pipeline}
\end{figure}

Some previous works\cite{schroff2015facenet, parkhi2015deep} successfully trained CNN models with the features being normalized in an end-to-end fashion. However, both of them used the triplet loss, which needs to sample triplets of face images during training. It is difficult to train because we usually need to implement hard mining algorithms to find non-trivial triplets\cite{schroff2015facenet}. Another route is to train a classification network using softmax loss\cite{sun2014deep, wu2015lightened} and regularizations to limit the intra-class variance\cite{wen2016discriminative, liu2016large}. Furthermore, some works combine the classification and metric learning loss functions together to train CNN models\cite{sun2014deep, Zhang2016Range}. All these methods that used classification loss functions, e.g. softmax loss, did not apply feature normalization, even though they all used normalized similarity measure, e.g. cosine similarity, to get the confidence of judging two samples being of the same identity at testing phase(Figure \ref{fig:pipeline}).

We did an experiment by normalizing both the features and the weights of the last inner-product layer to build a cosine layer in an ordinary CNN model. After sufficient iterations, the network still did not converge. After observing this phenomenon, we deeply dig into this problem. In this paper, we will find out the reason and propose methods to enable us to train the normalized features.


To sum up, in this work, we analyze and answer the questions mentioned above about the feature normalization and the model training:

\begin{myitemize}
\item Why is feature normalization so efficient when comparing the CNN features trained by classification loss, especially for softmax loss?

\item Why does directly optimizing the cosine similarity using softmax loss cause the network to fail to converge?

\item How to optimize a cosine similarity when using softmax loss?

\item Since models with softmax loss fail to converge after normalization, are there any other loss functions suitable for normalized features?
\end{myitemize}

For the first question, we explain it through a property of softmax loss in Section \ref{sec:necessity}. For the second and third questions, we provide a bound to describe the difficulty of using softmax loss to optimize a cosine similarity and propose using the scaled cosine similarity in Section \ref{sec:scale_softmax}. For the fourth question, we reformulate a set of loss functions in metric learning, such as contrastive loss and triplet loss to perform the classification task by introducing an `agent' strategy (Section \ref{sec:re_metric_learning}). Utilizing the `agent' strategy, there is no need to sample pairs and triplets of samples nor to implement the hard mining algorithm.

We also propose two tricks to improve performance for both static and video face verification. The first is to merge features extracted from both original image and mirror image by summation, while previous works usually merge the features by concatenation\cite{sun2014deep,wen2016discriminative}. The second is to use histogram of face similarities between video pairs instead of the mean\cite{parkhi2015deep,wen2016discriminative} or max\cite{xiang2016pose} similarity when making classification. 

Finally, by experiments, we show that normalization during training can promote the accuracies of two publicly available state-of-the-art models by $0.2 \sim 0.4\%$ on LFW\cite{huang2007labeled} and about $0.6\%$ on YTF\cite{wolf2011face}. 

\vspace{-2mm}
\section{Related Works}

\noindent \textbf{Normalization in Neural Network.} Normalization is a common operation in modern neural network models. Local Response Normalization and Local Contrast Normalization are studied in the AlexNet model\cite{krizhevsky2012imagenet}, even though these techniques are no longer common in modern models. Batch normalization\cite{ioffe2015batch} is widely used to accelerate the speed of neural network convergence by reducing the internal covariate shift of intermediate features. Weight normalization \cite{salimans2016weight} was proposed to normalize the weights of convolution layers and inner-product layers, and also lead to faster convergence speed. Layer normalization \cite{ba2016layer} tried to solve the batch size dependent problem of batch normalization, and works well on Recurrent Neural Networks.

\noindent \textbf{Face Verification.} Face verification is to decide whether two images containing faces represent the same person or two different people, and thus is important for access control or re-identification tasks. Face verification using deep learning techniques achieved a series of breakthroughs in recent years \cite{taigman2014deepface,lu2014surpassing,schroff2015facenet,parkhi2015deep,wen2016discriminative}. There are mainly two types of methods according to their loss functions. One type uses metric learning loss functions, such as contrastive loss\cite{chopra2005learning,yi2014learning} and triplet loss\cite{weinberger2009distance,schroff2015facenet, parkhi2015deep}. The other type uses softmax loss and treats the problem as a classification task, but also constrains the intra-class variance to get better generalization for comparing face features \cite{wen2016discriminative,liu2016large}. Some works also combine both kinds of loss functions\cite{yi2014learning,Zhang2016Range}. 

\noindent \textbf{Metric Learning.} Metric learning\cite{roweis2004neighbourhood,chopra2005learning,weinberger2009distance} tries to learn semantic distance measures and embeddings such that similar samples are nearer and different samples are further apart from each other on a manifold.
With the help of neural networks' enormous ability of representation learning, deep metric learning\cite{cai2012deep,LongSD15Fully} can do even better than the traditional methods. Recently, more complicated loss functions were proposed to get better local embedding structures\cite{oh2016deep,huang2016local,sohn2016improved}.

\noindent \textbf{Recent Works on Normalization.} Recently, cosine similarity \cite{liu2017learning} was used instead of the inner-product for training a CNN for person recognition, which is quite similar with face verification. The Cosine Loss proposed in \cite{liu2017learning} is quite similar with the one described in Section \ref{sec:scale_softmax}, normalizing both the features and weights. L2-softmax\cite{ranjan2017l2} shares a similar analysis about the convergence problem described in Section \ref{sec:scale_softmax}. In \cite{ranjan2017l2}, the authors also propose to add a scale parameter after normalization, but they only normalize the features. SphereFace\cite{liu2017sphereface} improves the performance of Large Margin Softmax\cite{liu2016large} by normalizing the weights of the last inner-product layer only. Von Mises-Fisher Mixture Model(vMFMM)\cite{hasnat2017von} interprets the hypersphere embedding as a mixture of von Mises-Fisher distributions. To sum up, the Cosine Loss\cite{liu2017learning}, vMFMM\cite{hasnat2017von} and our proposed loss functions optimize both features and weights, while the L2-softmax\cite{ranjan2017l2} normalizes the features only and the SphereFace\cite{liu2017sphereface} normalizes the weights only.

\section{$L_2$ Normalization Layer}

In this section, we answer the question why we should normalize the features when the loss function is softmax loss and why the network does not converge if we directly put a softmax loss on the normalized features.

\subsection{Necessity of Normalization}
\label{sec:necessity}

\begin{figure}
	\centering
	\includegraphics[scale=0.3]{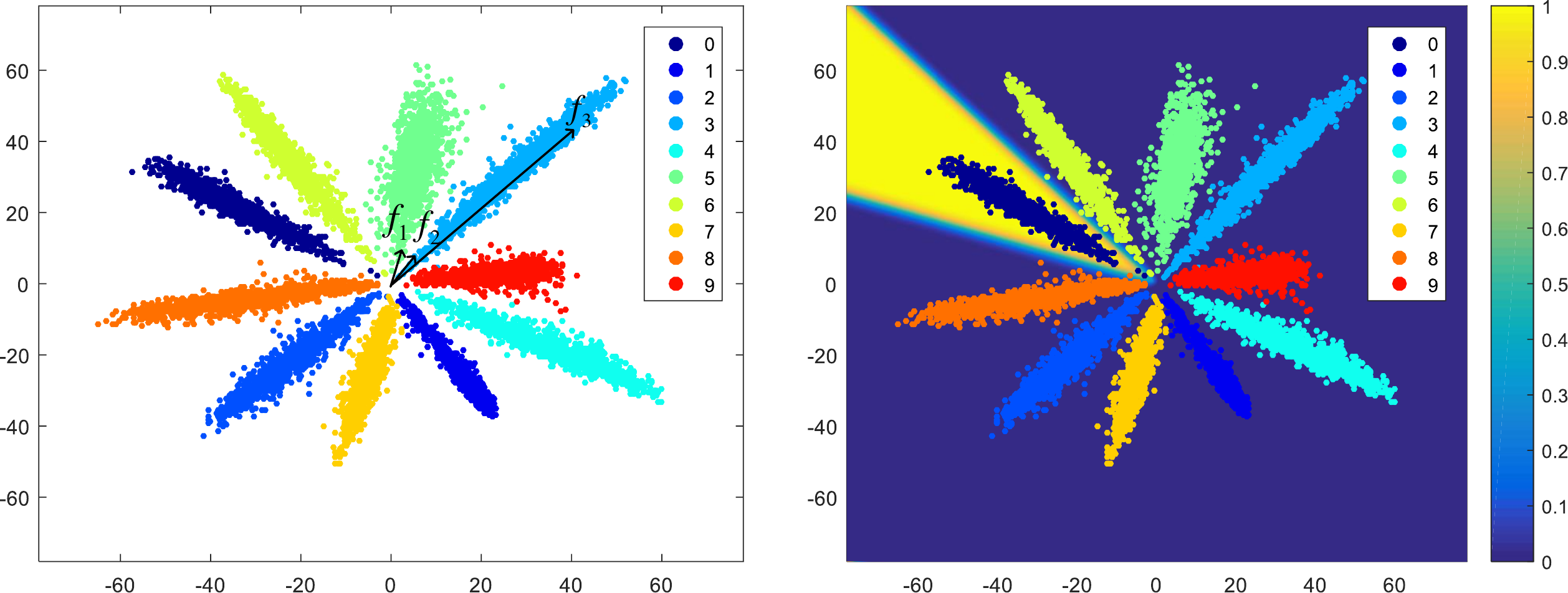}
	\caption{\emph{Left:} The optimized 2-dimensional feature distribution using softmax loss on MNIST\cite{lecun1998mnist} dataset. Note that the Euclidean distance between $\mathbf{f_1}$ and $\mathbf{f_2}$ is much smaller than the distance between $\mathbf{f_2}$ and $\mathbf{f_3}$, even though $\mathbf{f_2}$ and $\mathbf{f_3}$ are from the same class.  \emph{Right:} The softmax probability for class 0 on the 2-dimension plane. Best viewed in color. }
	\label{fig:mnist_dist_prob}
\end{figure}

In order to give an intuitive feeling about the softmax loss, we did a toy experiment of training a deeper LeNet\cite{lecun1998gradient} model on the MNIST dataset\cite{lecun1998mnist}. We reduced the number of the feature dimension to 2 and plot 10,000 2-dimensional features from the training set on a plane in Figure \ref{fig:mnist_dist_prob}. From the figure, we find that $\mathbf{f_2}$ can be much closer to $\mathbf{f_1}$ than to $\mathbf{f_3}$ if we use Euclidean distance as the metric. Hence directly using the features for comparison may lead to bad performance. At the same time, we find that the angles between feature vectors seem to be a good metric compared with Euclidean distance or inner-product operations. Actually, most previous work takes the cosine of the angle between feature vectors as the similarity \cite{sun2014deep,wu2015lightened,wen2016discriminative}, even though they all use softmax loss to train the network. Since the most common similarity metric for softmax loss is the inner-product with unnormalized features, there is a gap between the metrics used in the training and testing phases.

The reason why the softmax loss tends to create a `radial' feature distribution (Figure \ref{fig:mnist_dist_prob}) is that the softmax loss actually acts as the \emph{soft} version of \emph{max} operator. Scaling the feature vectors' magnitude does not affect the assignment of its class. Formally speaking, we recall the definition of the softmax loss,
\begin{equation}
\mathcal{L}_S = -\frac{1}{m}\sum_{i=1}^m{log\frac{e^{W_{y_i}^T \mathbf{f}_i +b_{y_i}}}{\sum_{j=1}^{n}{e^{W_j^T \mathbf{f}_i + b_j}}}},
\label{eq:softmax}
\end{equation}
where $m$ is the number of training samples, $n$ is the number of classes, $\mathbf{f}_i$ is the feature of the $i$-th sample, $y_i$ is the corresponding label in range $[1, n]$, $W$ and $b$ are the weight matrix and the bias vector of the last inner-product layer before the softmax loss, $W_j$ is the $j$-th column of $W$, which is corresponding to the $j$-th class. In the testing phase, we classify a sample by
\begin{equation}
Class(\mathbf{f}) = i = \arg\max_i{(W_i^T \mathbf{f} + b_i)}.
\end{equation}
In this case, we can infer that $(W_i \mathbf{f} + b_i) - (W_j \mathbf{f} + b_j) \ge 0, \forall j \in [1, n]$. Using this inequality, we obtain the following proposition.

\vspace{2mm}
\noindent \textbf{Proposition 1.} \emph{ For the softmax loss with \textbf{no-bias} inner-product similarity as its metric, let $P_i(\mathbf{f}) = \frac{e^{W_i^T \mathbf{f}}}{\sum_{j=1}^{n}{e^{W_j^T \mathbf{f}}}}$ denote the probability of $\mathbf{x}$ being classified as class $i$. For any given scale $s > 1$, if $i=\arg\max_j{(W_j^T \mathbf{f})}$, then $P_i(s\mathbf{f}) \ge P_i(\mathbf{f})$ always holds. }
\vspace{2mm}

The proof is given in Appendix \ref{sec:proof1}. This proposition implies that softmax loss always encourages well-separated features to have bigger magnitudes. This is the reason why the feature distribution of softmax is `radial'. However, we may not need this property as shown in Figure\ref{fig:mnist_dist_prob}. By normalization, we can eliminate its effect. Thus, we usually use the cosine of two feature vectors to measure the similarity of two samples.

\begin{figure}
	\centering
	\includegraphics[scale=0.3]{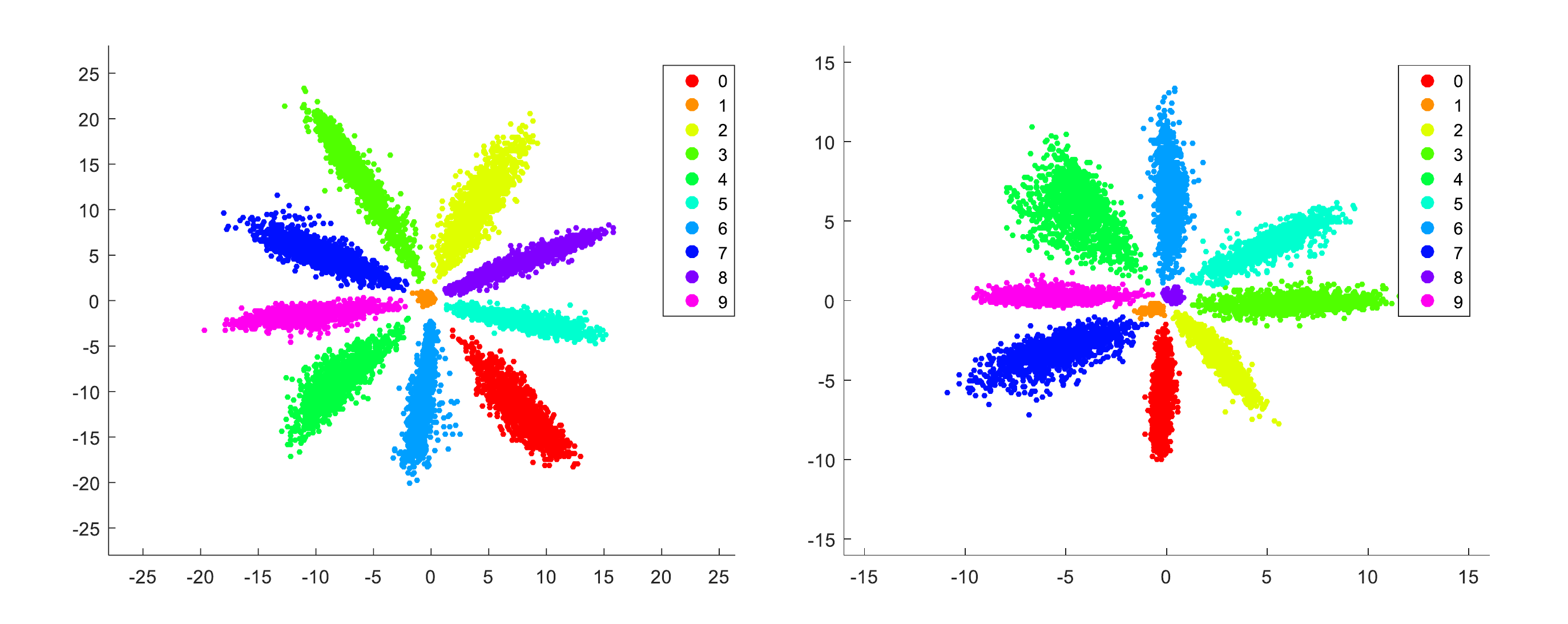}
	\caption{Two selected scatter diagrams when bias term is added after inner-product operation. Please note that there are one or two clusters that are located near the zero point. If we normalize the features of the center clusters, they would spread everywhere on the unit circle, which would cause misclassification. Best viewed in color. }
	\label{fig:mnist_dist_center}
\end{figure}

However, Proposition 1 does not hold if a bias term is added after the inner-product operation. In fact, the weight vector of the two classes could be the same and the model still could make a decision via the biases. We found this kind of case during the MNIST experiments and the scatters are shown in Figure \ref{fig:mnist_dist_center}. It can be discovered from the figure that the points of some classes all locate around the zero point, and after normalization the points from each of these classes may be spread out on the unit circle, overlapping with other classes. In these cases, feature normalization may destroy the discrimination ability of the specific classes. To avoid this kind of risk, we do not add the bias term before the softmax loss in this work, even though it is commonly used for classification tasks.

\subsection{Layer Definition}

In this paper, we define $\Vert \mathbf{x} \Vert_2 = \sqrt{\sum_i{\mathbf{x}_i^2}+\epsilon}$, where $\epsilon$ is a small positive value to prevent dividing zero. For an input vector $\mathbf{x} \in \mathcal{R}^n$, an $L_2$ normalization layer outputs the normalized vector,
\begin{equation}
\label{eq:normalize_x}
\mathbf{\tilde{x}} = \frac{\mathbf{x}}{\Vert \mathbf{x} \Vert_2} = \frac{\mathbf{x}}{\sqrt{\sum_i{\mathbf{x}_i^2}+\epsilon}}.
\end{equation}
Here $\mathbf{x}$ can be either the feature vector $\mathbf{f}$ or one column of the weight matrix $W_i$. In backward propagation, the gradient w.r.t. $\mathbf{x}$ can be obtained by the chain-rule,
\begin{equation}
\label{eq:norm_gradient}
\begin{aligned}
\frac{\partial \mathcal{L}}{\partial \mathbf{x}_i} & = \frac{\partial \mathcal{L}}{\partial \tilde{\mathbf{x}}_i}\frac{\partial \tilde{\mathbf{x}}_i}{\partial \mathbf{x}_i} + \sum_j{\frac{\partial \mathcal{L}}{\partial \tilde{\mathbf{x}}_j}\frac{\partial \tilde{\mathbf{x}}_j}{\partial \Vert \mathbf{x}  \Vert _2}\frac{\partial \Vert \mathbf{x}  \Vert _2}{\partial \mathbf{x}_i}}\\
& = 
\frac{\frac{\partial \mathcal{L}}{\partial \tilde{\mathbf{x}}_i} - 
	\tilde{\mathbf{x}}_i\sum_j{\frac{\partial \mathcal{L}}{\partial \tilde{\mathbf{x}}_j}\tilde{\mathbf{x}}_j}}
{\Vert \mathbf{x}  \Vert _2}.
\end{aligned}
\end{equation}

\begin{figure}
	\centering
	\includegraphics[scale=0.5]{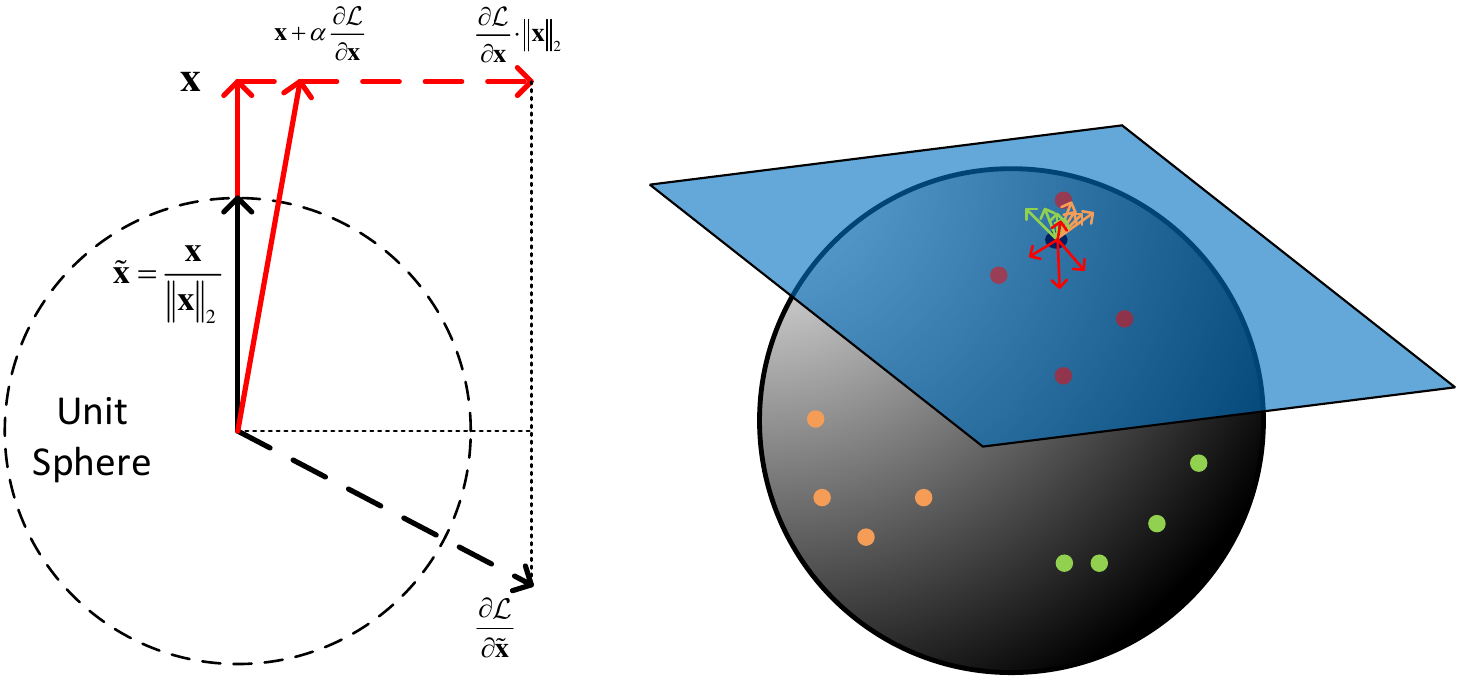}
	\caption{\emph{Left:} The normalization operation and its gradient in 2-dimensional space. Please note that $\|\mathbf{x}+\alpha \frac{\partial \mathcal{L}}{\partial \mathbf{x}}\|$ is always bigger than $\|\mathbf{x}\|$ for all $\alpha > 0$ because of the Pythagoras theorem. \emph{Right:} An example of the gradients w.r.t. the weight vector. All the gradients are in the tangent space of the unit sphere (denoted as the blue plane). The red, yellow and green points are normalized features from 3 different classes. The blue point is the normalized weight corresponding to the red class. Here we assume that the model tries to make features get close to their corresponding classes and away from other classes. Even though we illustrate the gradients applied on the normalized weight only, please note that opposite gradients are also applied on the normalized features (red, yellow, green points). Finally, all the gradients are accumulated together to decide which direction the weight should be updated. Best viewed in color, zoomed in.}
	\label{fig:geo_gradient}
\end{figure}

It is noteworthy that vector $\mathbf{x}$ and $\frac{\partial \mathcal{L}}{\partial \mathbf{x}}$ are orthogonal with each other, \emph{i.e.} $\langle \mathbf{x}, \frac{\partial \mathcal{L}}{\partial \mathbf{x}} \rangle = 0$. From a geometric perspective, the gradient $\frac{\partial \mathcal{L}}{\partial \mathbf{x}}$ is the projection of $\frac{\partial \mathcal{L}}{\partial \tilde{\mathbf{x}}}$ onto the tangent space of the unit hypersphere at normal vector $\mathbf{\tilde{x}}$ (see Figure \ref{fig:geo_gradient}). From Figure \ref{fig:geo_gradient} left, it can be inferred that after update, $\|\mathbf{x}\|_2$ always increases. In order to prevent $\|\mathbf{x}\|_2$ growing infinitely, weight decay is necessary on vector $\mathbf{x}$.

\subsection{Reformulating Softmax Loss}
\label{sec:scale_softmax}

Using the normalization layer, we can directly optimize the cosine similarity,
\begin{equation}
d(\mathbf{f}, \mathbf{W_i}) = \frac{\langle \mathbf{f}, \mathbf{W_i} \rangle}{\|\mathbf{f}\|_2\|\mathbf{W_i}\|_2},
\end{equation}
where $\mathbf{f}$ is the feature and $\mathbf{W_i}$ represents the $i$-th column of the weight matrix of the inner-product layer before softmax loss layer. However, after normalization, the network fails to converge.
The loss only decreases a little and then converges to a very big value within a few thousands of iterations. After that the loss does not decrease no matter how many iterations we train and how small the learning rate is.

This is mainly because the range of $d(\mathbf{f}, \mathbf{W_i})$ is only $[-1, 1]$ after normalization, while it is usually between $(-20, 20)$ and $(-80, 80)$ when we use an inner-product layer and softmax loss. This low range problem may prevent the probability $P_{y_i}(\mathbf{f};\mathbf{W}) = \frac{e^{\mathbf{W_{y_i}^T \mathbf{f}}}}{\sum_j^n{e^{\mathbf{W_j^T \mathbf{f}}}}}$, where $y_i$ is $\mathbf{f}$'s label, from getting close to $1$ even when the samples are well-separated. In the extreme case, $\frac{e^1}{e^1 + (n-1)e^{-1}}$ is very small ($0.45$ when $n=10$; $0.007$ when $n=1000$), even though in this condition the samples of all other classes are on the other side of the unit hypersphere. Since the gradient of softmax loss w.r.t. the ground truth label is $1-P_{y_i}$, the model will always try to give large gradients to the well separated samples, while the harder samples may not get sufficient gradients. 

To better understand this problem, we give a bound to clarify how small the softmax loss can be in the best case.

\vspace{2mm}
\noindent \textbf{Proposition 2.} (Softmax Loss Bound After Normalization)\emph{ Assume that every class has the same number of samples, and all the samples are well-separated, \emph{i.e.} each sample's feature is exactly same with its corresponding class's weight. If we normalize both the features and every column of the weights to have a norm of $\ell$, the softmax loss will have a lower bound, $\log \left( 1+ \left(n-1 \right)e^{-\frac{n}{n-1}\ell^2}\right)$, where $n$ is the class number.}
\vspace{2mm}

The proof is given in Appendix \ref{sec:proof2}. Even though reading the proof need patience, we still encourage readers to read it because you may get better understanding about the hypersphere manifold from it.

\begin{figure}
	\centering
	\includegraphics[scale=0.5]{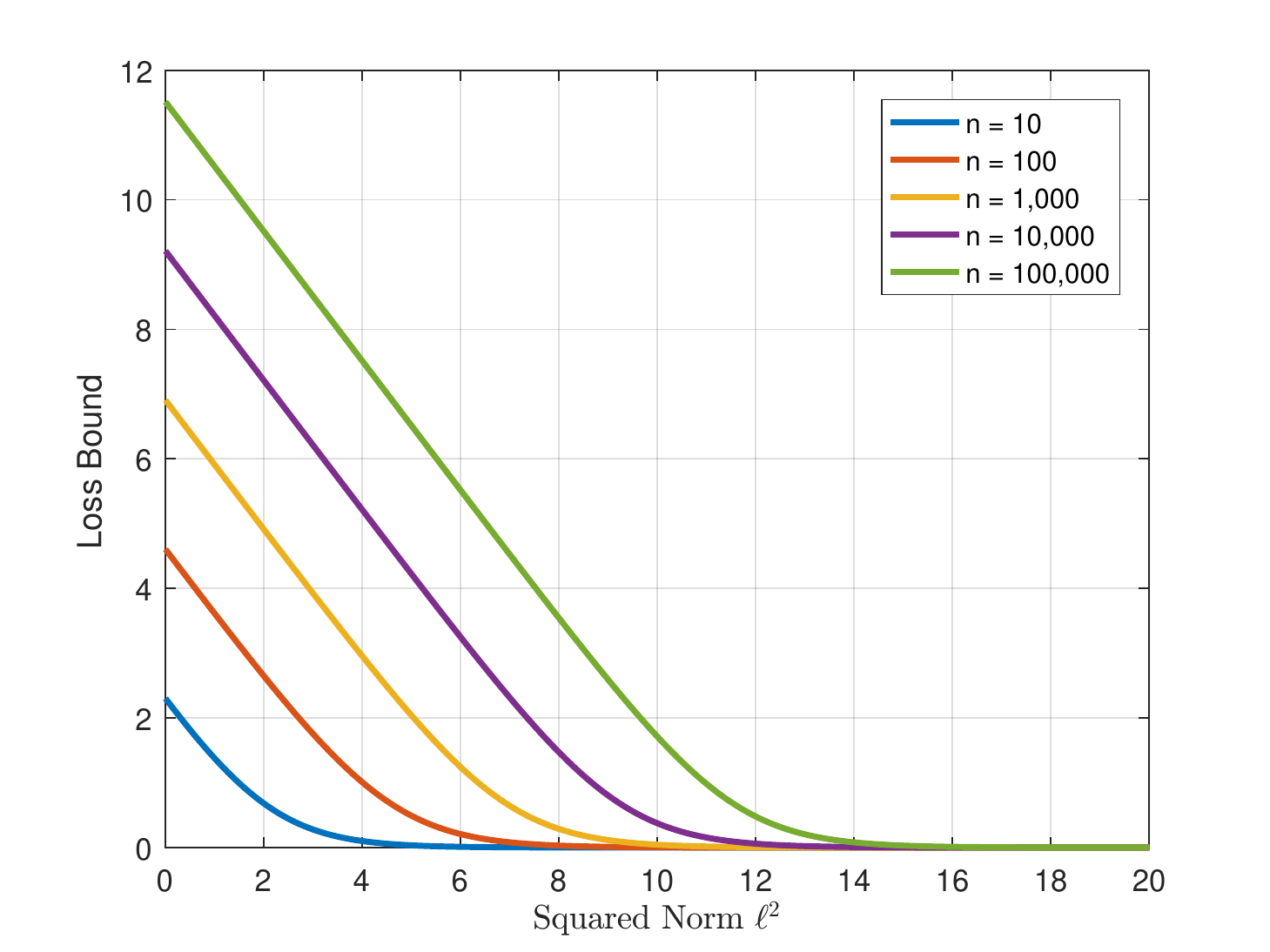}
	\caption{The softmax loss' lower bound as a function of features and weights' norm. Note that the $x$ axis is the squared norm $\ell^2$ because we add the scale parameter directly on the cosine distance in practice. }
	\label{fig:loss_bound}
\end{figure}

This bound implies that if we just normalize the features and weights to $1$, the softmax loss will be trapped at a very high value on training set, even if no regularization is applied. For a real example, if we train the model on the CASIA-Webface dataset ($n=10575$), the loss will decrease from about $9.27$ to about $8.50$. The bound for this condition is $8.27$, which is very close to the real value. This suggests that our bound is very tight. To give an intuition for the bound, we also plot the curve of the bound as a function of the norm $\ell$ in Figure \ref{fig:loss_bound}.

After we obtain the bound, the solution to the convergence problem is clear. By normalizing the features and columns of weight to a bigger value $\ell$ instead of 1, the softmax loss can continue to decrease. In practice, we may implement this by directly appending a scale layer after the cosine layer. The scale layer has only one learnable parameter $s=\ell^2$. We may also fix it to a value that is large enough referring to Figure \ref{fig:loss_bound}, say $20$ or $30$ for different class number. However, we prefer to make the parameter automatically learned by back-propagation instead of introducing a new hyper-parameter for elegance. Finally, the softmax loss with cosine distance is defined as
\begin{equation}
\mathcal{L}_\mathcal{S\prime}  = -\frac{1}{m}\sum_{i=1}^m{log\frac{e^{s\tilde{W}_{y_i}^T \tilde{\mathbf{f}}_i}}{\sum_{j=1}^{n}{e^{s\tilde{W}_j^T \tilde{\mathbf{f}}_i}}}},
\label{eq:normalize_softmax}
\end{equation}
where $\tilde{\mathbf{x}}$ is the normalized $\mathbf{x}$.

\section{Reformulating Metric Learning}
\label{sec:re_metric_learning}


Metric Learning, or specifically deep metric learning in this work, usually takes pairs or triplets of samples as input, and outputs the distance between them. In deep metric models, it is a common strategy to normalize the final features\cite{schroff2015facenet,parkhi2015deep,oh2016deep}. It seems that normalization does not cause any problems for metric learning loss functions. However, metric learning is more difficult to train than classification because the possible input pairs or triplets in metric learning models are very large, namely $\mathcal{O}(N^2)$ combinations for pairs and $\mathcal{O}(N^3)$ combinations for triplets, where $N$ is the amount of training samples. It is almost impossible to deal with all possible combinations during training, so sampling and hard mining algorithms are usually necessary\cite{schroff2015facenet}, which are tricky and time-consuming. By contrast, in a classification task, we usually feed the data iteratively into the model, namely the input data is in order of $\mathcal{O}(N)$. In this section, we attempt to reformulate some metric learning loss functions to do the classification task, while keeping their compatibility with the normalized features.

\begin{figure}
	\centering
	\includegraphics[scale=0.35]{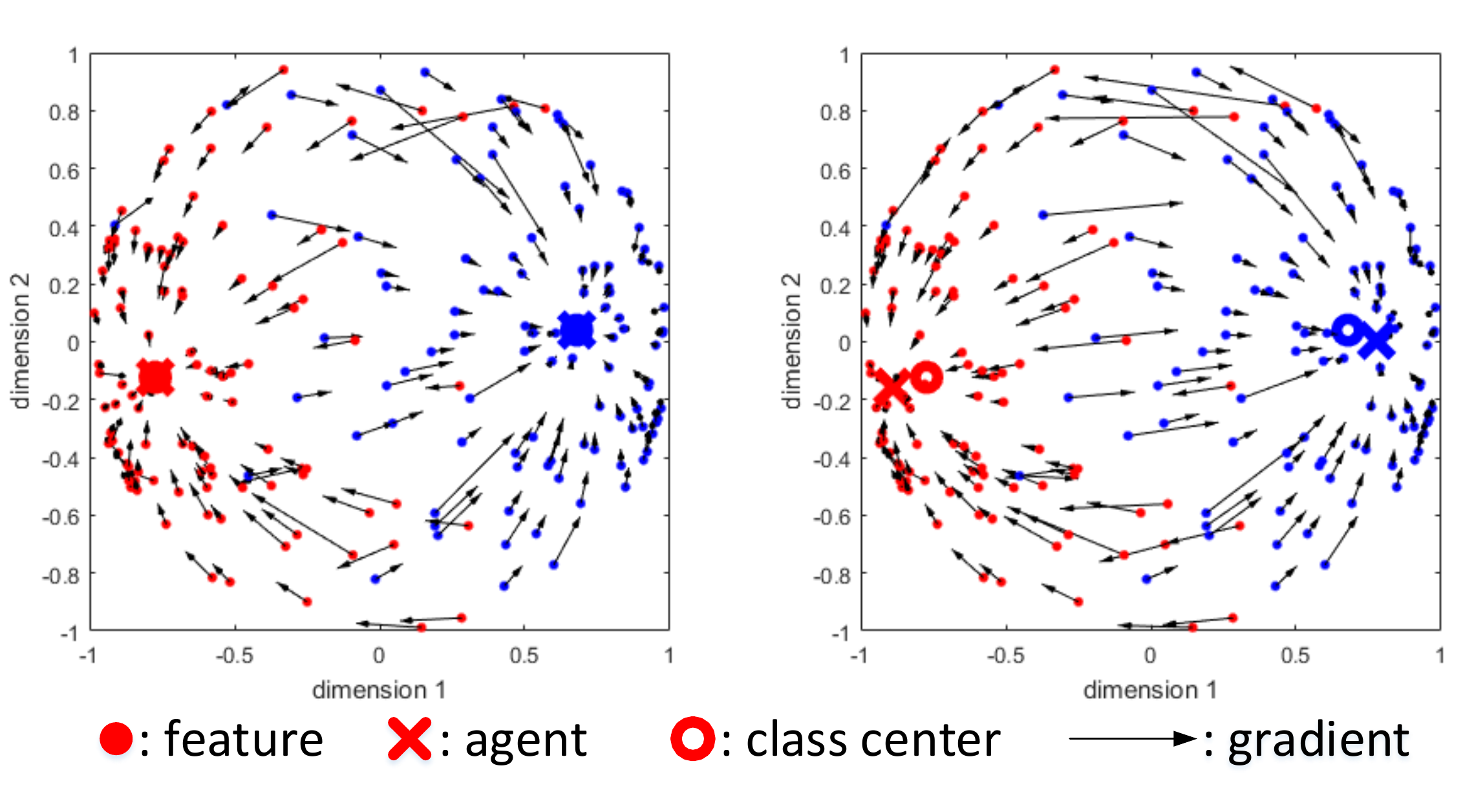}
	\caption{ Illustration of how the C-contrastive loss works with two classes on a $3$-d sphere (projected on a $2$-d plane). \emph{Left:} The special case of $m=0$. In this case, the agents are only influenced by features from their own classes. The agents will finally converge to the centers of their corresponding classes. \emph{Right:} Normal case of $m=1$. In this case, the agents are influenced by all the features in the same classes and other classes' features in the margin. Hence the agents are shifted away from the boundary of the two classes. The features will follow their agents through the intra-class term $\|\tilde{\mathbf{f}}_i - \tilde{W}_j\|_2^2, c_i = j$ as the gradients shown in the figure. Best viewed in color.}
	\label{fig:agent}
\end{figure}

The most widely used metric learning methods in the face verification community are the contrastive loss\cite{sun2014deep,yi2014learning},
\begin{equation}
\mathcal{L}_\mathcal{C} = \left\{
{\begin{array}{*{20}{l}}
	{\|\tilde{\mathbf{f}}_i - \tilde{\mathbf{f}}_j\|_2^2,\quad c_i = c_j}\\
	{\max(0, m - \|\tilde{\mathbf{f}}_i - \tilde{\mathbf{f}}_j\|_2^2),\quad c_i \neq c_j}
	\end{array} 
} \right. ,
\label{eq:contrastive}
\end{equation}
and the triplet loss\cite{schroff2015facenet, parkhi2015deep},
\begin{equation}
\mathcal{L}_\mathcal{T} = max(0, m + \|\tilde{\mathbf{f}}_i - \tilde{\mathbf{f}}_j\|_2^2 - \|\tilde{\mathbf{f}}_i - \tilde{\mathbf{f}}_k\|_2^2), \quad c_i = c_j, c_i \neq c_k ,
\label{eq:triplet}
\end{equation}
where the two $m$'s are the margins. 
Both of the two loss functions optimize the normalized Euclidean distance between feature pairs. Note that after normalization, the reformulated softmax loss can also be seen as optimizing the normalized Euclidean distance,
\begin{equation}
\begin{aligned}
\mathcal{L}_\mathcal{S\prime} & = -\frac{1}{m}\sum_{i=1}^m{log\frac{e^{s\tilde{W}_{y_i}^T \tilde{\mathbf{f}_i}}}{\sum_{j=1}^{n}{e^{s\tilde{W}_j^T \tilde{\mathbf{f}_i}}}}}\\
& = -\frac{1}{m}\sum_{i=1}^m{log\frac{e^{-\frac{s}{2}\|\tilde{\mathbf{f}_i} - \tilde{W}_{y_i} \|_2^2}}{\sum_{j=1}^{n}{e^{-\frac{s}{2}\|\tilde{\mathbf{f}_i} - \tilde{W}_j\|_2^2 }}}},
\end{aligned}
\label{eq:normalize_softmax_l2}
\end{equation}
because $\|\tilde{\mathbf{x}} - \tilde{\mathbf{y}}\|_2^2 = 2 - 2\tilde{\mathbf{x}}^T \tilde{\mathbf{y}}$. Inspired by this formulation, we modify one of the features to be one column of a weight matrix $W \in \mathbb{R}^{d\times n}$, where $d$ is the dimension of the feature and $n$ is the class number. We call column $W_i$ as the `agent' of the $i$-th class. The weight matrix $W$ can be learned through back-propagation just as the inner-product layer. In this way, we can get a classification version of the contrastive loss,
\begin{equation}
\mathcal{L}_\mathcal{C\prime} = \left\{
{\begin{array}{*{20}{l}}
	{\|\tilde{\mathbf{f}}_i - \tilde{W}_j\|_2^2,\quad c_i = j}\\
	{\max(0, m - \|\tilde{\mathbf{f}}_i - \tilde{W}_j\|_2^2),\quad c_i \neq j}
	\end{array} 
} \right. ,
\label{eq:contrastive_class}
\end{equation}
and the triplet loss,
\begin{equation}
\mathcal{L}_\mathcal{T\prime} = max(0, m + \|\tilde{\mathbf{f}}_i - \tilde{W}_j\|_2^2 - \|\tilde{\mathbf{f}}_i - \tilde{W}_k\|_2^2), \quad c_i = j, c_i \neq k .
\label{eq:triplet_class}
\end{equation}

To distinguish these two loss functions from their metric learning versions, we call them \emph{C-contrastive loss} and \emph{C-triplet loss} respectively, denoting that these loss functions are designed for classification. 

Intuitively, $W_j$ acts as a summarizer of the features in $j$-th class. If all classes are well-separated by the margin, the $W_j$'s will roughly correspond to the means of features in each class (Figure \ref{fig:agent} left). In more complicated tasks, features of different classes may be overlapped with each other. Then the $W_j$'s will be shifted away from the boundaries. The marginal features (hard examples) are guided to have bigger gradients in this case (Figure \ref{fig:agent} right), which means they move further than easier samples during update.

\begin{figure}
	\centering
	\includegraphics[scale=0.25]{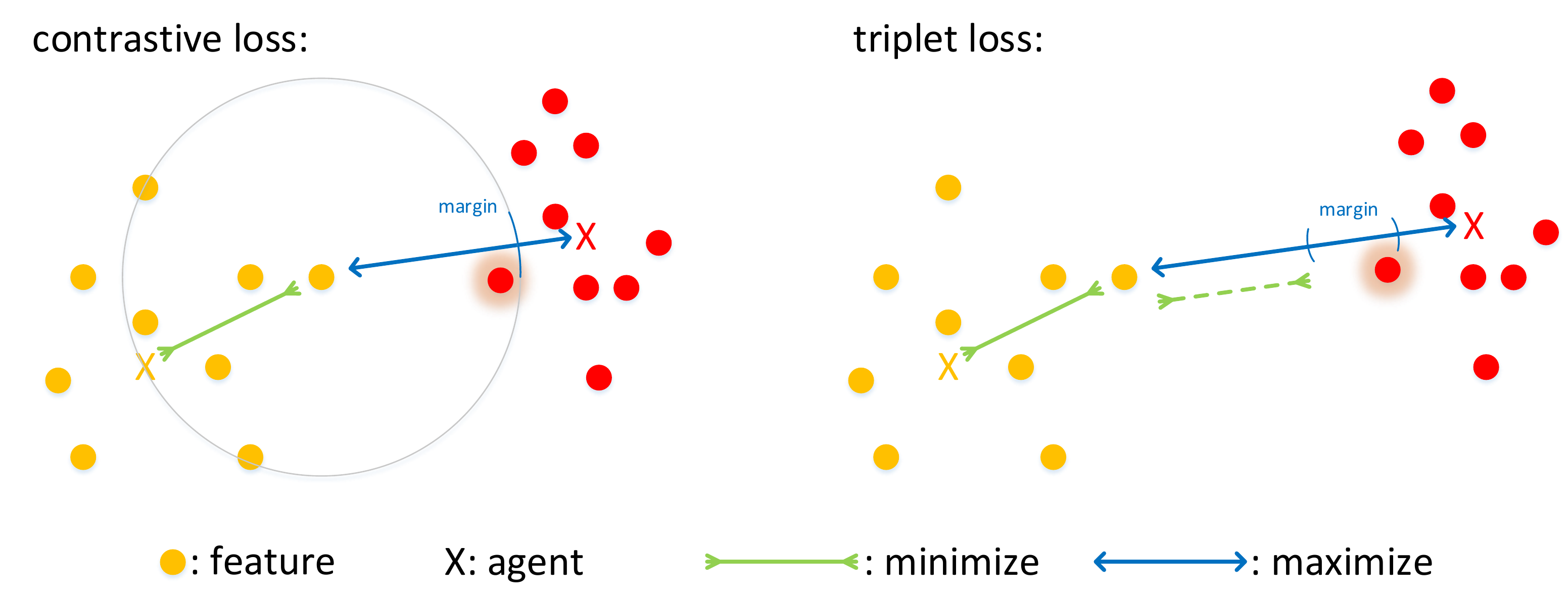}
	\caption{ Classification version of contrastive loss (Left) and triplet loss (Right). The shadowed points are the marginal features that got omitted due to the `agent' strategy. In the original version of the two losses, the shadowed points are also optimized. Best viewed in color. }
	\label{fig:class_contrastive_triplet}
\end{figure}

However, there are some side effect of the agent strategy. After reformulation, some of the marginal features may not be optimized if we still use the same margin as the original version (Figure \ref{fig:class_contrastive_triplet}). Thus, we need larger margins to make more features get optimized. Mathematically, the error caused by the agent approximation is given by the following proposition.

\vspace{2mm}
\noindent \textbf{Proposition 3.} \emph{ Using an agent for each class instead of a specific sample would cause a distortion of $\frac{1}{n_{\mathcal{C}_i}}\sum_{j \in \mathcal{C}_i}{\left(d(f_0, f_j) - d(f_0, W_i)\right)^2}$, where $W_i$ is the agent of the $i$th-class. The distortion is bounded by $\frac{1}{n_{\mathcal{C}_i}}\sum_{j \in \mathcal{C}_i}{d(f_j, W_i)^2}$.}
\vspace{2mm}

The proof is given in Appendix \ref{sec:proof3}. This bound gives us a theoretical guidance of setting the margins. We can compute it on-the-fly during training using moving-average and display it to get better feelings about the progress. Empirically, the bound $\frac{1}{n_{\mathcal{C}_i}}\sum_{j \in \mathcal{C}_i}{d(f_j, W_i)^2}$ is usually $0.5\sim 0.6$. The recommendation value of the margins of the modified contrastive loss and triplet loss is $1$ and $0.8$ respectively.

Note that setting the margin used to be complicated work\cite{yi2014learning}. Following their work, we have to suspend training and search for a new margin for every several epochs. However, we no longer need to perform such a searching algorithm after applying normalization. Through normalization, the scale of features' magnitude is fixed, which makes it possible to fix the margin, too. In this strategy, we will not try to train models using the C-contrastive loss or the C-triplet loss without normalization because this is difficult. 

\section{Experiment}
\label{sec:experiment}
In this section, we first describe the experiment settings in Section \ref{sec:implementation_details}. Then we evaluate our method on two different datasets with two different models in Section \ref{sec:lfw} and \ref{sec:ytf}. Codes and models are released at \url{https://github.com/happynear/NormFace}.
\subsection{Implementation Details}
\label{sec:implementation_details}
\noindent \textbf{Baseline works.} To verify our algorithm's universality, we choose two works as our baseline, Wu \emph{et. al.}'s model \cite{wu2015lightened}\footnote{\url{https://github.com/AlfredXiangWu/face_verification_experiment}} (Wu's model, for short) and Wen \emph{et. al.}'s model \cite{wen2016discriminative}\footnote{\url{https://github.com/ydwen/caffe-face}} (Wen's model, for short). Wu's model is a 10-layer plain CNN with Maxout\cite{goodfellow2013maxout} activation unit. Wen's model is a 28-layer ResNet\cite{he2016deep} trained with both softmax loss and center loss. Neither of these two models apply feature normalization or weight normalization. We strictly follow all the experimental settings as their papers, including the datasets\footnote{ Since the identity label of the Celebrity+\cite{liu2015deep} dataset is not publicly available, we follow Wen's released model which is trained on CASIA-Webface \cite{yi2014learning} only. Wu's model is also trained on CASIA-Webface \cite{yi2014learning} only. }, the image resolution, the pre-processing methods and the evaluation criteria.

\noindent \textbf{Training.} The proposed loss functions are appended after the feature layer, \emph{i.e.} the second last inner-product layer. The features and columns of weight matrix are normalized to make their $L_2$ norm to be $1$. Then the features and columns of the weight matrix are sent into a pairwise distance layer, \emph{i.e.} inner-product layer to produce a cosine similarity or Euclidean distance layer to produce a normalized Euclidean distance. After calculating all the similarities or distances between each feature and each column, the proposed loss functions will give the final loss and gradients to the distances. The whole network models are trained end to end. To speed up the training procedure, we fine-tune the networks from the baseline models. Thus, a relatively small learning rate, say 1e-4 for Wu's model and 1e-3 for Wen's model, are applied to update the network through stochastic gradient descent (SGD) with momentum of $0.9$.

\noindent \textbf{Evaluation.} Two datasets are utilized to evaluate the performance, one is Labeled Face in the Wild (LFW)\cite{huang2007labeled} and another is Youtube Face (YTF)\cite{wolf2011face}. $10$-fold validation is used to evaluate the performance for both datasets. After the training models converge, we continue to train them for $5,000$ iterations\footnote{In each iteration we train 256 samples, i.e. the batch size is 256.}, during which we save a snapshot for every $1,000$ iterations. Then we run the evaluation codes on the five saved snapshots separately and calculate an average score to reduce disturbance. We extract features from both the frontal face and its mirror image and merge the two features by \emph{element-wise summation}. Principle Component Analysis (PCA) is then applied on the training subset of the evaluation dataset to fit the features to the target domain. Similarity score is computed by the cosine distance of two sample's features after PCA. All the evaluations are based on the similarity scores of image pairs.

\begin{figure}
	\centering
	\includegraphics[scale=0.5]{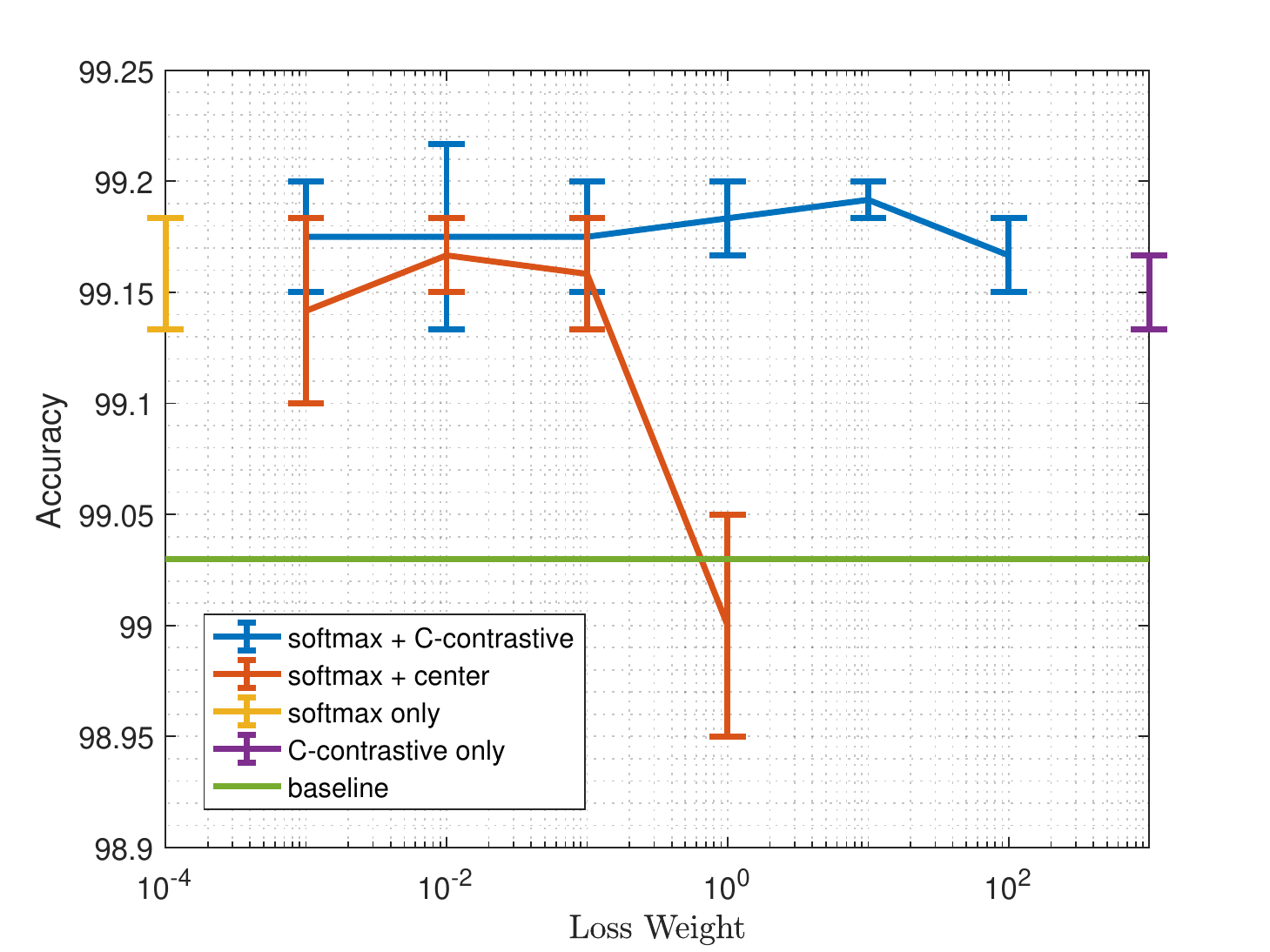}
	\caption{ LFW accuracies as a function of the loss weight of C-contrastive loss or center loss with error bars. All these methods use the normalization strategy except for the baseline. }
	\label{fig:loss_weight}
\end{figure}

\subsection{Experiments on LFW} 
\label{sec:lfw}

The LFW dataset\cite{huang2007labeled} contains $13,233$ images from $5,749$ identities, with large variations in pose, expression and illumination. All the images are collected from Internet. We evaluate our methods through two different protocols on LFW, one is the standard \emph{unrestricted with labeled outside data} \cite{huang2014labeled}, which is evaluated on $6,000$ image pairs, and another is \emph{BLUFR} \cite{liao2014benchmark} which utilize all $13,233$ images. It is noteworthy that there are three same identities in CASIA-Webface\cite{yi2014learning} and LFW\cite{huang2007labeled}. We delete them during training to build a complete open-set validation.

\begin{table}[!htb]
	\caption{Results on LFW 6,000 pairs using Wen's model\cite{wen2016discriminative}}
	\label{tab:loss_standard_lfw}
	\begin{tabular}{ccc}
		\toprule
		loss function & Normalization & Accuracy\\
		\midrule
		softmax & No & 98.28\%\\
		softmax + dropout & No & 98.35\%\\
		softmax + center\cite{wen2016discriminative} & No & 99.03\%\\
		softmax & feature only & 98.72\% \\
		softmax & weight only & 98.95\% \\
		softmax & Yes & 99.16\% $\pm$ 0.025\%\\
		softmax + center &  Yes & 99.17\% $\pm$ 0.017\%\\
		C-contrasitve & Yes & 99.15\% $\pm$ 0.017\%\\
		C-triplet & Yes & 99.11\% $\pm$ 0.008\%\\
		C-triplet + center & Yes & 99.13\% $\pm$ 0.017\%\\
		softmax + C-contrastive & Yes & 99.19\% $\pm$ 0.008\%\\
		\bottomrule
	\end{tabular}
\end{table}

\begin{table*}
	\caption{Results on LFW BLUFR\cite{liao2014benchmark} protocol}
	\label{tab:loss_blufr}
	\begin{tabular}{ccccc}
		\toprule
		model & loss function & Normalization & TPR@FAR=0.1\% & DIR@FAR=1\%\\
		\midrule
		ResNet & softmax + center\cite{wen2016discriminative} & No & 93.35\% & 67.86\%\\
		ResNet & softmax & Yes & 95.77\% & 73.92\%\\
		ResNet & C-triplet + center & Yes & 95.73\% & 76.12\%\\
		ResNet & softmax + C-contrastive & Yes & 95.83\% & 77.18\%\\
		\midrule
		MaxOut & softmax\cite{wu2015lightened} & No & 89.12\% & 61.79\%\\
		MaxOut & softmax & Yes & 90.64\% & 65.22\%\\
		MaxOut & C-contrastive & Yes & 90.32\% & 68.14\%\\
		\bottomrule
	\end{tabular}
\end{table*}

We carefully test almost all combinations of the loss functions on the standard \emph{unrestricted with labeled outside data} protocol. The results are listed in Table \ref{tab:loss_standard_lfw}. Cosine similarity is used by \emph{softmax + any} loss functions. The distance used by \emph{C-contrastive} and \emph{C-triplet} loss is the squared normalized Euclidean distance. The \emph{C-triplet + center} loss is implemented by forcing to optimize $\|\mathbf{x}_i - W_j\|_2^2$ even if $m + \|\mathbf{x}_i - W_j\|_2^2 - \|\mathbf{x}_i - W_k\|_2^2$ is less than $0$. From Table \ref{tab:loss_standard_lfw} we can conclude that the loss functions have minor influence on the accuracy, and the normalization is the key factor to promote the performance. When combining the softmax loss with the C-contrastive loss or center loss, we need to add a hyper-parameter to make balance between the two losses. The highest accuracy, $\mathbf{99.2167\%}$, is obtained by softmax + $0.01\ *$ C-contrastive. However, pure softmax with normalization already works reasonably well. 

We have also designed two ablation experiments of normalizing the features only or normalizing the columns of weight matrix only. 
During experiments we find that the scale parameter is necessary when normalizing the feature, while normalizing the weight does not need it. We cannot explain it so far. This is tricky but the network will collapse if the scale parameter is not properly added. 
From Table \ref{tab:loss_standard_lfw} we can conclude that normalizing the feature causes performance degradation, while normalizing the weight has little influence on the accuracy. Note that these two special cases of softmax loss are also fine-tuned based on Wen's model. When training from scratch, normalizing the weights only will cause the network collapse, while normalizing the features only will lead to a worse accuracy, $98.45\%$, which is better than the conventional softmax loss, but much worse than state-of-the-art loss functions.

\begin{table}[!htb]
	\caption{Results on LFW 6,000 pairs using Wu's model\cite{wu2015lightened}}
	\label{tab:loss_standard_lfw_wu}
	\begin{tabular}{ccc}
		\toprule
		loss function & Normalization & Accuracy\\
		\midrule
		softmax & No & 98.13\%\\
		softmax + mirror & No & 98.41\%\\
		softmax & Yes & 98.75\% $\pm$ 0.008\%\\
		C-contrastive & Yes & 98.78\% $\pm$ 0.017\%\\
		softmax + C-contrastive & Yes & 98.71\% $\pm$ 0.017\%\\
		\bottomrule
	\end{tabular}
\end{table}

In Figure \ref{fig:loss_weight}, we show the effect of the loss weights when using two loss functions. As shown in the figure, the C-contrastive loss is more robust to the loss weight. This is not surprising because C-contrastive loss can train a model by itself only, while the center loss, which only optimizes the intra-class variance, should be trained with other supervised losses together. 

To make our experiment more convincing, we also train some of the loss functions on Wu's model\cite{wu2015lightened}. The results are listed in Table \ref{tab:loss_standard_lfw_wu}. Note that in \cite{wu2015lightened}, Wu et. al. did not perform face mirroring when they evaluated their methods. In Table \ref{tab:loss_standard_lfw_wu}, we also present the result of their model after face mirroring and feature merging. As is shown in the table, the normalization operation still gives a significant boost to the performance.

On BLUFR protocol, the normalization technique works even better. Here we only compare some of the models with the baseline (Table \ref{tab:loss_blufr}). From Table \ref{tab:loss_blufr} we can see that normalization could boost the performance significantly, which reveals that normalization technique could perform much better when the false alarm rate (FAR) is low.

\begin{figure}
	\centering
	\includegraphics[scale=0.4]{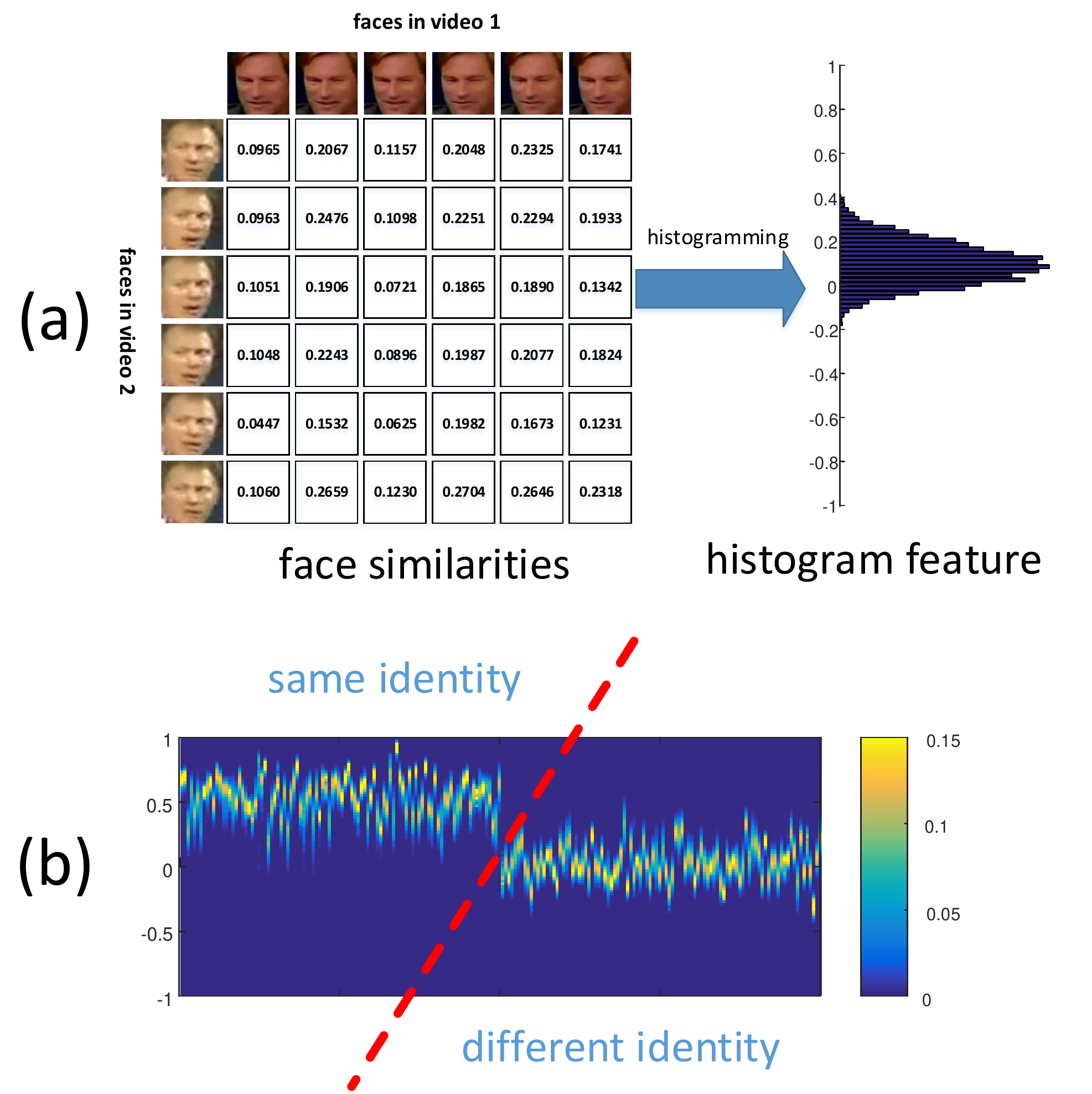}
	\caption{ (a): Illustration of how to generate a histogram feature for a pair of videos. We firstly create a pairwise score matrix by computing the cosine similarity between two face images from different video sequences. Then we accumulate all the scores in the matrix to create a histogram. (b): Visualization of histogram features extracted from $200$ video pairs with both same identities and different identities. After collecting all histogram features, support vector machine (SVM) using histogram intersection kernel(HIK) is utilized to make a binary classification. }
	\label{fig:videoface}
\end{figure}

\subsection{Experiments on YTF}
\label{sec:ytf}

The YTF dataset\cite{wolf2011face} consists of 3,425 videos of 1,595 different people, with an
average of 2.15 videos per person. We follow the \emph{unrestricted	with labeled outside data} protocol, which takes $5,000$ video pairs to evaluate the performance.

Previous works usually extract face features from all frames or some selected frames in a video. Then two videos can construct a confidence matrix $C$ in which each element $C_{ij}$ is the cosine distance of face features extracted from the $i$-th frame of the first video and $j$-th frame of the second video. The final score is computed by the average of all all elements in $C$. The one dimension score is then used to train a classifier, say SVM, to get the threshold of same identity or different identity.

Here we propose to use the histogram of elements in $C$ as the feature to train the classifier. The bin of the histogram is set to $100$ (Figure \ref{fig:videoface}(a)). Then SVM with histogram intersection kernel (HIK-SVM)\cite{barla2003histogram} is utilized to make a two-class classification (Figure \ref{fig:videoface}(b)). This method encodes more information compared to the one dimensional mean value, and leads to better performance on video face verification.

\begin{table}
	\caption{Results on YTF with Wen's model\cite{wen2016discriminative}}
	\label{tab:ytf}
	\begin{tabular}{ccc}
		\toprule
		loss function & Normalization & Accuracy\\
		\midrule
		softmax + center\cite{wen2016discriminative} & No & 93.74\%\\
		softmax & Yes & 94.24\%\\
		softmax + HIK-SVM & Yes & 94.56\%\\
		C-triplet + center & Yes & 94.3\%\\
		C-triplet + center + HIK-SVM & Yes & 94.58\%\\
		softmax + C-contrastive & Yes & 94.34\%\\
		softmax + C-contrastive + HIK-SVM & Yes & 94.72\%\\
		\bottomrule
	\end{tabular}
\end{table}

The results are listed in Table \ref{tab:ytf}. The models that perform better on LFW also show superior performance on YTF. Moreover, the newly proposed score histogram technique (HIK-SVM in the table) can improve the accuracy further by a significant gap.

\section{Conclusion and Future Work}

In this paper, we propose to apply $L_2$ normalization operation on the features and the weight of the last inner-product layer when training a classification model. We explain the necessity of the normalization operation from both analytic and geometric perspective. Two kinds of loss functions are proposed to effectively train the normalized feature. One is a reformulated softmax loss with a scale layer inserted between the cosine score and the loss. Another is designed inspired by metric learning. We introduce an agent strategy to avoid the need of hard sample mining, which is a tricky and time-consuming work. Experiments on two different models both show superior performance over models without normalization. From three theoretical propositions, we also provide some guidance on the hyper-parameter setting, such as the bias term (Proposition 1), the scale parameter (Proposition 2) and the margin (Proposition 3).


Currently we can only \emph{fine-tune} the network with normalization techniques based on other models. If we train a model with C-contrastive loss function, the final result is just as good as center loss\cite{wen2016discriminative}. But if we fine-tune a model, either Wen's model\cite{wen2016discriminative} or Wu's model\cite{wu2015lightened}, the performance could be further improved as shown in Table \ref{tab:loss_standard_lfw} and Table \ref{tab:loss_standard_lfw_wu}. More efforts are needed to find a way to train a model from scratch, while preserving at least a similar performance as fine-tuning.

Our methods and analysis in this paper are general. They can be used in other metric learning tasks, such as person re-identification or image retrieval. We will apply the proposed methods on these tasks in the future.

\section{Acknowledgement}

This paper is funded by Office of Naval Research (N00014-15-1-2356), National Science Foundation (CCF-1317376), the National Natural Science Foundation of China (61671125,
61201271, 61301269) and the State Key Laboratory of Synthetical Automation for Process
Industries (NO. PAL-N201401). 

We thank Chenxu Luo and Hao Zhu for their assistance in formula derivation.

\bibliographystyle{ACM-Reference-Format}
\balance
\bibliography{sigproc} 


\begin{thebibliography}{00}


\ifx \showCODEN    \undefined \def \showCODEN     #1{\unskip}     \fi
\ifx \showDOI      \undefined \def \showDOI       #1{{\tt DOI:}\penalty0{#1}\ }
  \fi
\ifx \showISBNx    \undefined \def \showISBNx     #1{\unskip}     \fi
\ifx \showISBNxiii \undefined \def \showISBNxiii  #1{\unskip}     \fi
\ifx \showISSN     \undefined \def \showISSN      #1{\unskip}     \fi
\ifx \showLCCN     \undefined \def \showLCCN      #1{\unskip}     \fi
\ifx \shownote     \undefined \def \shownote      #1{#1}          \fi
\ifx \showarticletitle \undefined \def \showarticletitle #1{#1}   \fi
\ifx \showURL      \undefined \def \showURL       #1{#1}          \fi
\providecommand\bibfield[2]{#2}
\providecommand\bibinfo[2]{#2}
\providecommand\natexlab[1]{#1}
\providecommand\showeprint[2][]{arXiv:#2}

\bibitem[\protect\citeauthoryear{Ba, Kiros, and Hinton}{Ba
  et~al\mbox{.}}{2016}]%
        {ba2016layer}
\bibfield{author}{\bibinfo{person}{Jimmy~Lei Ba}, \bibinfo{person}{Jamie~Ryan
  Kiros}, {and} \bibinfo{person}{Geoffrey~E Hinton}.}
  \bibinfo{year}{2016}\natexlab{}.
\newblock \showarticletitle{Layer normalization}.
\newblock \bibinfo{journal}{{\em arXiv preprint arXiv:1607.06450\/}}
  (\bibinfo{year}{2016}).
\newblock


\bibitem[\protect\citeauthoryear{Barla, Odone, and Verri}{Barla
  et~al\mbox{.}}{2003}]%
        {barla2003histogram}
\bibfield{author}{\bibinfo{person}{Annalisa Barla}, \bibinfo{person}{Francesca
  Odone}, {and} \bibinfo{person}{Alessandro Verri}.}
  \bibinfo{year}{2003}\natexlab{}.
\newblock \showarticletitle{Histogram intersection kernel for image
  classification}. In \bibinfo{booktitle}{{\em Image Processing, 2003. ICIP
  2003. Proceedings. 2003 International Conference on}},
  Vol.~\bibinfo{volume}{3}. IEEE, \bibinfo{pages}{III--513}.
\newblock


\bibitem[\protect\citeauthoryear{Cai, Wang, Xiao, Chen, and Zhou}{Cai
  et~al\mbox{.}}{2012}]%
        {cai2012deep}
\bibfield{author}{\bibinfo{person}{Xinyuan Cai}, \bibinfo{person}{Chunheng
  Wang}, \bibinfo{person}{Baihua Xiao}, \bibinfo{person}{Xue Chen}, {and}
  \bibinfo{person}{Ji Zhou}.} \bibinfo{year}{2012}\natexlab{}.
\newblock \showarticletitle{Deep nonlinear metric learning with independent
  subspace analysis for face verification}. In \bibinfo{booktitle}{{\em ACM
  international conference on Multimedia}}. ACM, \bibinfo{pages}{749--752}.
\newblock


\bibitem[\protect\citeauthoryear{Chopra, Hadsell, and LeCun}{Chopra
  et~al\mbox{.}}{2005}]%
        {chopra2005learning}
\bibfield{author}{\bibinfo{person}{Sumit Chopra}, \bibinfo{person}{Raia
  Hadsell}, {and} \bibinfo{person}{Yann LeCun}.}
  \bibinfo{year}{2005}\natexlab{}.
\newblock \showarticletitle{Learning a similarity metric discriminatively, with
  application to face verification}. In \bibinfo{booktitle}{{\em {IEEE}
  Conference on Computer Vision and Pattern Recognition}},
  Vol.~\bibinfo{volume}{1}. IEEE, \bibinfo{pages}{539--546}.
\newblock


\bibitem[\protect\citeauthoryear{Girshick, Donahue, Darrell, and
  Malik}{Girshick et~al\mbox{.}}{2014}]%
        {girshick2014rich}
\bibfield{author}{\bibinfo{person}{Ross Girshick}, \bibinfo{person}{Jeff
  Donahue}, \bibinfo{person}{Trevor Darrell}, {and} \bibinfo{person}{Jagannath
  Malik}.} \bibinfo{year}{2014}\natexlab{}.
\newblock \showarticletitle{Rich feature hierarchies for accurate object
  detection and semantic segmentation}. In \bibinfo{booktitle}{{\em {IEEE}
  Conference on Computer Vision and Pattern Recognition}}.
  \bibinfo{pages}{580--587}.
\newblock


\bibitem[\protect\citeauthoryear{Goodfellow, Warde-Farley, Mirza, Courville,
  and Bengio}{Goodfellow et~al\mbox{.}}{2013}]%
        {goodfellow2013maxout}
\bibfield{author}{\bibinfo{person}{Ian~J Goodfellow}, \bibinfo{person}{David
  Warde-Farley}, \bibinfo{person}{Mehdi Mirza}, \bibinfo{person}{Aaron~C
  Courville}, {and} \bibinfo{person}{Yoshua Bengio}.}
  \bibinfo{year}{2013}\natexlab{}.
\newblock \showarticletitle{Maxout Networks.}
\newblock \bibinfo{journal}{{\em International Conference on Machine
  Learning\/}}  \bibinfo{volume}{28} (\bibinfo{year}{2013}),
  \bibinfo{pages}{1319--1327}.
\newblock


\bibitem[\protect\citeauthoryear{He, Zhang, Ren, and Sun}{He
  et~al\mbox{.}}{2016}]%
        {he2016deep}
\bibfield{author}{\bibinfo{person}{Kaiming He}, \bibinfo{person}{Xiangyu
  Zhang}, \bibinfo{person}{Shaoqing Ren}, {and} \bibinfo{person}{Jian Sun}.}
  \bibinfo{year}{2016}\natexlab{}.
\newblock \showarticletitle{Deep residual learning for image recognition}. In
  \bibinfo{booktitle}{{\em {IEEE} Conference on Computer Vision and Pattern
  Recognition}}. \bibinfo{pages}{770--778}.
\newblock


\bibitem[\protect\citeauthoryear{Huang, Loy, and Tang}{Huang
  et~al\mbox{.}}{2016}]%
        {huang2016local}
\bibfield{author}{\bibinfo{person}{Chen Huang}, \bibinfo{person}{Chen~Change
  Loy}, {and} \bibinfo{person}{Xiaoou Tang}.} \bibinfo{year}{2016}\natexlab{}.
\newblock \showarticletitle{Local similarity-aware deep feature embedding}. In
  \bibinfo{booktitle}{{\em Advances in Neural Information Processing Systems}}.
  \bibinfo{pages}{1262--1270}.
\newblock


\bibitem[\protect\citeauthoryear{Huang and Learned-Miller}{Huang and
  Learned-Miller}{2014}]%
        {huang2014labeled}
\bibfield{author}{\bibinfo{person}{Gary~B Huang} {and} \bibinfo{person}{Erik
  Learned-Miller}.} \bibinfo{year}{2014}\natexlab{}.
\newblock \showarticletitle{Labeled faces in the wild: Updates and new
  reporting procedures}.
\newblock \bibinfo{journal}{{\em Dept. Comput. Sci., Univ. Massachusetts
  Amherst, Amherst, MA, USA, Tech. Rep\/}} (\bibinfo{year}{2014}),
  \bibinfo{pages}{14--003}.
\newblock


\bibitem[\protect\citeauthoryear{Huang, Ramesh, Berg, and Learned-Miller}{Huang
  et~al\mbox{.}}{2007}]%
        {huang2007labeled}
\bibfield{author}{\bibinfo{person}{Gary~B Huang}, \bibinfo{person}{Manu
  Ramesh}, \bibinfo{person}{Tamara Berg}, {and} \bibinfo{person}{Erik
  Learned-Miller}.} \bibinfo{year}{2007}\natexlab{}.
\newblock \bibinfo{booktitle}{{\em Labeled faces in the wild: A database for
  studying face recognition in unconstrained environments}}.
\newblock \bibinfo{type}{{T}echnical {R}eport}. \bibinfo{institution}{Technical
  Report 07-49, University of Massachusetts, Amherst}.
\newblock


\bibitem[\protect\citeauthoryear{Ioffe and Szegedy}{Ioffe and Szegedy}{2015}]%
        {ioffe2015batch}
\bibfield{author}{\bibinfo{person}{Sergey Ioffe} {and}
  \bibinfo{person}{Christian Szegedy}.} \bibinfo{year}{2015}\natexlab{}.
\newblock \showarticletitle{Batch normalization: Accelerating deep network
  training by reducing internal covariate shift}.
\newblock \bibinfo{journal}{{\em arXiv preprint arXiv:1502.03167\/}}
  (\bibinfo{year}{2015}).
\newblock


\bibitem[\protect\citeauthoryear{Krizhevsky, Sutskever, and Hinton}{Krizhevsky
  et~al\mbox{.}}{2012}]%
        {krizhevsky2012imagenet}
\bibfield{author}{\bibinfo{person}{Alex Krizhevsky}, \bibinfo{person}{Ilya
  Sutskever}, {and} \bibinfo{person}{Geoffrey~E Hinton}.}
  \bibinfo{year}{2012}\natexlab{}.
\newblock \showarticletitle{Imagenet classification with deep convolutional
  neural networks}. In \bibinfo{booktitle}{{\em Advances in neural information
  processing systems}}. \bibinfo{pages}{1097--1105}.
\newblock


\bibitem[\protect\citeauthoryear{LeCun, Bottou, Bengio, and Haffner}{LeCun
  et~al\mbox{.}}{1998a}]%
        {lecun1998gradient}
\bibfield{author}{\bibinfo{person}{Yann LeCun}, \bibinfo{person}{L{\'e}on
  Bottou}, \bibinfo{person}{Yoshua Bengio}, {and} \bibinfo{person}{Patrick
  Haffner}.} \bibinfo{year}{1998}\natexlab{a}.
\newblock \showarticletitle{Gradient-based learning applied to document
  recognition}.
\newblock \bibinfo{journal}{{\it Proc. IEEE}} \bibinfo{volume}{86},
  \bibinfo{number}{11} (\bibinfo{year}{1998}), \bibinfo{pages}{2278--2324}.
\newblock


\bibitem[\protect\citeauthoryear{LeCun, Cortes, and Burges}{LeCun
  et~al\mbox{.}}{1998b}]%
        {lecun1998mnist}
\bibfield{author}{\bibinfo{person}{Yann LeCun}, \bibinfo{person}{Corinna
  Cortes}, {and} \bibinfo{person}{Christopher Burges}.}
  \bibinfo{year}{1998}\natexlab{b}.
\newblock \bibinfo{title}{The mnist database of handwritten digits}.
\newblock   (\bibinfo{year}{1998}).
\newblock
\showURL{%
\url{http://yann.lecun.com/exdb/mnist/}}


\bibitem[\protect\citeauthoryear{Liao, Lei, Yi, and Li}{Liao
  et~al\mbox{.}}{2014}]%
        {liao2014benchmark}
\bibfield{author}{\bibinfo{person}{Shengcai Liao}, \bibinfo{person}{Zhen Lei},
  \bibinfo{person}{Dong Yi}, {and} \bibinfo{person}{Stan~Z Li}.}
  \bibinfo{year}{2014}\natexlab{}.
\newblock \showarticletitle{A benchmark study of large-scale unconstrained face
  recognition}. In \bibinfo{booktitle}{{\em IEEE International Joint Conference
  on Biometrics}}. IEEE, \bibinfo{pages}{1--8}.
\newblock


\bibitem[\protect\citeauthoryear{Liu, Wen, Yu, and Yang}{Liu
  et~al\mbox{.}}{2016}]%
        {liu2016large}
\bibfield{author}{\bibinfo{person}{Weiyang Liu}, \bibinfo{person}{Yandong Wen},
  \bibinfo{person}{Zhiding Yu}, {and} \bibinfo{person}{Meng Yang}.}
  \bibinfo{year}{2016}\natexlab{}.
\newblock \showarticletitle{Large-Margin Softmax Loss for Convolutional Neural
  Networks}. In \bibinfo{booktitle}{{\em International Conference on Machine
  Learning}}. \bibinfo{pages}{507--516}.
\newblock


\bibitem[\protect\citeauthoryear{Liu, Li, and Wang}{Liu et~al\mbox{.}}{2017}]%
        {liu2017learning}
\bibfield{author}{\bibinfo{person}{Yu Liu}, \bibinfo{person}{Hongyang Li},
  {and} \bibinfo{person}{Xiaogang Wang}.} \bibinfo{year}{2017}\natexlab{}.
\newblock \showarticletitle{Learning Deep Features via Congenerous Cosine Loss
  for Person Recognition}.
\newblock \bibinfo{journal}{{\em arXiv preprint arXiv:1702.06890\/}}
  (\bibinfo{year}{2017}).
\newblock


\bibitem[\protect\citeauthoryear{Liu, Luo, Wang, and Tang}{Liu
  et~al\mbox{.}}{2015}]%
        {liu2015deep}
\bibfield{author}{\bibinfo{person}{Ziwei Liu}, \bibinfo{person}{Ping Luo},
  \bibinfo{person}{Xiaogang Wang}, {and} \bibinfo{person}{Xiaoou Tang}.}
  \bibinfo{year}{2015}\natexlab{}.
\newblock \showarticletitle{Deep learning face attributes in the wild}. In
  \bibinfo{booktitle}{{\em Proceedings of the IEEE International Conference on
  Computer Vision}}. \bibinfo{pages}{3730--3738}.
\newblock


\bibitem[\protect\citeauthoryear{Long, Shelhamer, and Darrell}{Long
  et~al\mbox{.}}{2015}]%
        {LongSD15Fully}
\bibfield{author}{\bibinfo{person}{Jonathan Long}, \bibinfo{person}{Evan
  Shelhamer}, {and} \bibinfo{person}{Trevor Darrell}.}
  \bibinfo{year}{2015}\natexlab{}.
\newblock \showarticletitle{Fully convolutional networks for semantic
  segmentation}. In \bibinfo{booktitle}{{\em {IEEE} Conference on Computer
  Vision and Pattern Recognition}}. \bibinfo{pages}{3431--3440}.
\newblock


\bibitem[\protect\citeauthoryear{Lu and Tang}{Lu and Tang}{2014}]%
        {lu2014surpassing}
\bibfield{author}{\bibinfo{person}{Chaochao Lu} {and} \bibinfo{person}{Xiaoou
  Tang}.} \bibinfo{year}{2014}\natexlab{}.
\newblock \showarticletitle{Surpassing human-level face verification
  performance on LFW with GaussianFace}.
\newblock \bibinfo{journal}{{\em arXiv preprint arXiv:1404.3840\/}}
  (\bibinfo{year}{2014}).
\newblock


\bibitem[\protect\citeauthoryear{Md. Abul~Hasnat}{Md. Abul~Hasnat}{2017}]%
        {hasnat2017von}
\bibfield{author}{\bibinfo{person}{Jonathan Milgram Stéphane Gentric
  Liming~Chen Md. Abul~Hasnat, Julien~Bohné}.}
  \bibinfo{year}{2017}\natexlab{}.
\newblock \showarticletitle{von Mises-Fisher Mixture Model-based Deep learning:
  Application to Face Verification}.
\newblock \bibinfo{journal}{{\em arXiv preprint arXiv:1706.04264\/}}
  (\bibinfo{year}{2017}).
\newblock


\bibitem[\protect\citeauthoryear{Oh~Song, Xiang, Jegelka, and Savarese}{Oh~Song
  et~al\mbox{.}}{2016}]%
        {oh2016deep}
\bibfield{author}{\bibinfo{person}{Hyun Oh~Song}, \bibinfo{person}{Yu Xiang},
  \bibinfo{person}{Stefanie Jegelka}, {and} \bibinfo{person}{Silvio Savarese}.}
  \bibinfo{year}{2016}\natexlab{}.
\newblock \showarticletitle{Deep metric learning via lifted structured feature
  embedding}. In \bibinfo{booktitle}{{\em {IEEE} Conference on Computer Vision
  and Pattern Recognition}}. \bibinfo{pages}{4004--4012}.
\newblock


\bibitem[\protect\citeauthoryear{Parkhi, Vedaldi, and Zisserman}{Parkhi
  et~al\mbox{.}}{2015}]%
        {parkhi2015deep}
\bibfield{author}{\bibinfo{person}{Omkar~M Parkhi}, \bibinfo{person}{Andrea
  Vedaldi}, {and} \bibinfo{person}{Andrew Zisserman}.}
  \bibinfo{year}{2015}\natexlab{}.
\newblock \showarticletitle{Deep Face Recognition.}. In
  \bibinfo{booktitle}{{\em BMVC}}, Vol.~\bibinfo{volume}{1}.
  \bibinfo{pages}{6}.
\newblock


\bibitem[\protect\citeauthoryear{Ranjan, Castillo, and Chellappa}{Ranjan
  et~al\mbox{.}}{2017}]%
        {ranjan2017l2}
\bibfield{author}{\bibinfo{person}{Rajeev Ranjan}, \bibinfo{person}{Carlos~D.
  Castillo}, {and} \bibinfo{person}{Rama Chellappa}.}
  \bibinfo{year}{2017}\natexlab{}.
\newblock \showarticletitle{L2-constrained Softmax Loss for Discriminative Face
  Verification}.
\newblock \bibinfo{journal}{{\em arXiv preprint arXiv:1703.09507\/}}
  (\bibinfo{year}{2017}).
\newblock


\bibitem[\protect\citeauthoryear{Roweis, Hinton, and Salakhutdinov}{Roweis
  et~al\mbox{.}}{2004}]%
        {roweis2004neighbourhood}
\bibfield{author}{\bibinfo{person}{Sam Roweis}, \bibinfo{person}{Geoffrey
  Hinton}, {and} \bibinfo{person}{Ruslan Salakhutdinov}.}
  \bibinfo{year}{2004}\natexlab{}.
\newblock \showarticletitle{Neighbourhood component analysis}.
\newblock \bibinfo{journal}{{\em Advances in Neural Information Processing
  Systems\/}}  \bibinfo{volume}{17} (\bibinfo{year}{2004}),
  \bibinfo{pages}{513--520}.
\newblock


\bibitem[\protect\citeauthoryear{Rudin et~al\mbox{.}}{Rudin
  et~al\mbox{.}}{1964}]%
        {rudin1964principles}
\bibfield{author}{\bibinfo{person}{Walter Rudin} {and}
  \bibinfo{person}{others}.} \bibinfo{year}{1964}\natexlab{}.
\newblock \bibinfo{booktitle}{{\em Principles of mathematical analysis, Chapter
  10}}. Vol.~\bibinfo{volume}{3}.
\newblock \bibinfo{publisher}{McGraw-Hill New York}.
\newblock


\bibitem[\protect\citeauthoryear{Salimans and Kingma}{Salimans and
  Kingma}{2016}]%
        {salimans2016weight}
\bibfield{author}{\bibinfo{person}{Tim Salimans} {and}
  \bibinfo{person}{Diederik~P Kingma}.} \bibinfo{year}{2016}\natexlab{}.
\newblock \showarticletitle{Weight normalization: A simple reparameterization
  to accelerate training of deep neural networks}. In \bibinfo{booktitle}{{\em
  Advances in Neural Information Processing Systems}}.
  \bibinfo{pages}{901--901}.
\newblock


\bibitem[\protect\citeauthoryear{Schroff, Kalenichenko, and Philbin}{Schroff
  et~al\mbox{.}}{2015}]%
        {schroff2015facenet}
\bibfield{author}{\bibinfo{person}{Florian Schroff}, \bibinfo{person}{Dmitry
  Kalenichenko}, {and} \bibinfo{person}{James Philbin}.}
  \bibinfo{year}{2015}\natexlab{}.
\newblock \showarticletitle{Facenet: A unified embedding for face recognition
  and clustering}. In \bibinfo{booktitle}{{\em {IEEE} Conference on Computer
  Vision and Pattern Recognition}}. \bibinfo{pages}{815--823}.
\newblock


\bibitem[\protect\citeauthoryear{Simonyan and Zisserman}{Simonyan and
  Zisserman}{2014}]%
        {simonyan2014very}
\bibfield{author}{\bibinfo{person}{Karen Simonyan} {and}
  \bibinfo{person}{Andrew Zisserman}.} \bibinfo{year}{2014}\natexlab{}.
\newblock \showarticletitle{Very Deep Convolutional Networks for Large-Scale
  Image Recognition}.
\newblock \bibinfo{journal}{{\em arXiv preprint arXiv:1409.1556\/}}
  (\bibinfo{year}{2014}).
\newblock


\bibitem[\protect\citeauthoryear{Sohn}{Sohn}{2016}]%
        {sohn2016improved}
\bibfield{author}{\bibinfo{person}{Kihyuk Sohn}.}
  \bibinfo{year}{2016}\natexlab{}.
\newblock \showarticletitle{Improved deep metric learning with multi-class
  n-pair loss objective}. In \bibinfo{booktitle}{{\em Advances in Neural
  Information Processing Systems}}. \bibinfo{pages}{1849--1857}.
\newblock


\bibitem[\protect\citeauthoryear{Sun, Chen, Wang, and Tang}{Sun
  et~al\mbox{.}}{2014}]%
        {sun2014deep}
\bibfield{author}{\bibinfo{person}{Yi Sun}, \bibinfo{person}{Yuheng Chen},
  \bibinfo{person}{Xiaogang Wang}, {and} \bibinfo{person}{Xiaoou Tang}.}
  \bibinfo{year}{2014}\natexlab{}.
\newblock \showarticletitle{Deep learning face representation by joint
  identification-verification}. In \bibinfo{booktitle}{{\em Advances in neural
  information processing systems}}. \bibinfo{pages}{1988--1996}.
\newblock


\bibitem[\protect\citeauthoryear{Szegedy, Liu, Jia, Sermanet, Reed, Anguelov,
  Erhan, Vanhoucke, and Rabinovich}{Szegedy et~al\mbox{.}}{2015}]%
        {szegedy2014going}
\bibfield{author}{\bibinfo{person}{Christian Szegedy}, \bibinfo{person}{Wei
  Liu}, \bibinfo{person}{Yangqing Jia}, \bibinfo{person}{Pierre Sermanet},
  \bibinfo{person}{Scott Reed}, \bibinfo{person}{Dragomir Anguelov},
  \bibinfo{person}{Dumitru Erhan}, \bibinfo{person}{Vincent Vanhoucke}, {and}
  \bibinfo{person}{Andrew Rabinovich}.} \bibinfo{year}{2015}\natexlab{}.
\newblock \showarticletitle{Going deeper with convolutions}. In
  \bibinfo{booktitle}{{\em {IEEE} Conference on Computer Vision and Pattern
  Recognition}}. \bibinfo{pages}{1--9}.
\newblock


\bibitem[\protect\citeauthoryear{Taigman, Yang, Ranzato, and Wolf}{Taigman
  et~al\mbox{.}}{2014}]%
        {taigman2014deepface}
\bibfield{author}{\bibinfo{person}{Yaniv Taigman}, \bibinfo{person}{Ming Yang},
  \bibinfo{person}{Marc'Aurelio Ranzato}, {and} \bibinfo{person}{Lior Wolf}.}
  \bibinfo{year}{2014}\natexlab{}.
\newblock \showarticletitle{Deepface: Closing the gap to human-level
  performance in face verification}. In \bibinfo{booktitle}{{\em {IEEE}
  Conference on Computer Vision and Pattern Recognition}}.
  \bibinfo{pages}{1701--1708}.
\newblock


\bibitem[\protect\citeauthoryear{Weinberger and Saul}{Weinberger and
  Saul}{2009}]%
        {weinberger2009distance}
\bibfield{author}{\bibinfo{person}{Kilian~Q Weinberger} {and}
  \bibinfo{person}{Lawrence~K Saul}.} \bibinfo{year}{2009}\natexlab{}.
\newblock \showarticletitle{Distance metric learning for large margin nearest
  neighbor classification}.
\newblock \bibinfo{journal}{{\em Journal of Machine Learning Research\/}}
  \bibinfo{volume}{10}, \bibinfo{number}{Feb} (\bibinfo{year}{2009}),
  \bibinfo{pages}{207--244}.
\newblock


\bibitem[\protect\citeauthoryear{Weiyang~Liu and Song}{Weiyang~Liu and
  Song}{2017}]%
        {liu2017sphereface}
\bibfield{author}{\bibinfo{person}{Zhiding Yu Ming Li Bhiksha~Raj Weiyang~Liu,
  Yandong~Wen} {and} \bibinfo{person}{Le Song}.}
  \bibinfo{year}{2017}\natexlab{}.
\newblock \showarticletitle{SphereFace: Deep Hypersphere Embedding for Face
  Recognition}. In \bibinfo{booktitle}{{\em Proceedings of the IEEE conference
  on computer vision and pattern recognition}}.
\newblock


\bibitem[\protect\citeauthoryear{Wen, Zhang, Li, and Qiao}{Wen
  et~al\mbox{.}}{2016}]%
        {wen2016discriminative}
\bibfield{author}{\bibinfo{person}{Yandong Wen}, \bibinfo{person}{Kaipeng
  Zhang}, \bibinfo{person}{Zhifeng Li}, {and} \bibinfo{person}{Yu Qiao}.}
  \bibinfo{year}{2016}\natexlab{}.
\newblock \showarticletitle{A Discriminative Feature Learning Approach for Deep
  Face Recognition}. In \bibinfo{booktitle}{{\em European Conference on
  Computer Vision}}. Springer, \bibinfo{pages}{499--515}.
\newblock


\bibitem[\protect\citeauthoryear{Wolf, Hassner, and Maoz}{Wolf
  et~al\mbox{.}}{2011}]%
        {wolf2011face}
\bibfield{author}{\bibinfo{person}{Lior Wolf}, \bibinfo{person}{Tal Hassner},
  {and} \bibinfo{person}{Itay Maoz}.} \bibinfo{year}{2011}\natexlab{}.
\newblock \showarticletitle{Face recognition in unconstrained videos with
  matched background similarity}. In \bibinfo{booktitle}{{\em {IEEE} Conference
  on Computer Vision and Pattern Recognition}}. IEEE,
  \bibinfo{pages}{529--534}.
\newblock


\bibitem[\protect\citeauthoryear{Wu, He, and Sun}{Wu et~al\mbox{.}}{2015}]%
        {wu2015lightened}
\bibfield{author}{\bibinfo{person}{Xiang Wu}, \bibinfo{person}{Ran He}, {and}
  \bibinfo{person}{Zhenan Sun}.} \bibinfo{year}{2015}\natexlab{}.
\newblock \showarticletitle{A Lightened CNN for Deep Face Representation}.
\newblock \bibinfo{journal}{{\em arXiv preprint arXiv:1511.02683\/}}
  (\bibinfo{year}{2015}).
\newblock


\bibitem[\protect\citeauthoryear{Xiang and Tran}{Xiang and Tran}{2016}]%
        {xiang2016pose}
\bibfield{author}{\bibinfo{person}{Xiang Xiang} {and} \bibinfo{person}{Trac~D
  Tran}.} \bibinfo{year}{2016}\natexlab{}.
\newblock \showarticletitle{Pose-Selective Max Pooling for Measuring
  Similarity}.
\newblock \bibinfo{journal}{{\em Lecture Notes in Computer Science\/}}
  \bibinfo{volume}{10165} (\bibinfo{year}{2016}).
\newblock


\bibitem[\protect\citeauthoryear{Yi, Lei, Liao, and Li}{Yi
  et~al\mbox{.}}{2014}]%
        {yi2014learning}
\bibfield{author}{\bibinfo{person}{Dong Yi}, \bibinfo{person}{Zhen Lei},
  \bibinfo{person}{Shengcai Liao}, {and} \bibinfo{person}{Stan~Z Li}.}
  \bibinfo{year}{2014}\natexlab{}.
\newblock \showarticletitle{Learning face representation from scratch}.
\newblock \bibinfo{journal}{{\em arXiv preprint arXiv:1411.7923\/}}
  (\bibinfo{year}{2014}).
\newblock


\bibitem[\protect\citeauthoryear{Zhang, Fang, Wen, Li, and Qiao}{Zhang
  et~al\mbox{.}}{2016}]%
        {Zhang2016Range}
\bibfield{author}{\bibinfo{person}{Xiao Zhang}, \bibinfo{person}{Zhiyuan Fang},
  \bibinfo{person}{Yandong Wen}, \bibinfo{person}{Zhifeng Li}, {and}
  \bibinfo{person}{Yu Qiao}.} \bibinfo{year}{2016}\natexlab{}.
\newblock \showarticletitle{Range Loss for Deep Face Recognition with
  Long-tail}.
\newblock \bibinfo{journal}{{\em arXiv preprint arXiv:1611.08976\/}}
  (\bibinfo{year}{2016}).
\newblock


\end{thebibliography}
\clearpage
\section{Appendix}
\subsection{Proof of Proposition 1}
\label{sec:proof1}
\noindent \textbf{Proposition 1.} \emph{ For the softmax loss with \textbf{no-bias} inner-product similarity as its metric, let $P_i(\mathbf{f}) = \frac{e^{W_i^T \mathbf{f}}}{\sum_{j=1}^{n}{e^{W_j^T \mathbf{f}}}}$ denote the probability of $\mathbf{f}$ being classified as class $i$. For a given scale $s > 1$, if $i=\arg\max_j{(W_j^T \mathbf{f})}$, then $P_i(s\mathbf{f}) \ge P_i(\mathbf{f})$ always holds. }

\noindent \textbf{Proof:} Let $t = s -1$, after scaling, we have,
\begin{equation}
\begin{aligned}
P_i\left({s\mathbf{f}}\right) & = \frac{e^{W_i^T [(1+t)\mathbf{f}]}}{\sum_{j=1}^{n}{e^{W_j^T [(1+t)\mathbf{f}]}}} \\
& = \frac{e^{W_i^T \mathbf{f}}}{\sum_{j=1}^{n}{e^{W_j^T \mathbf{f} + t (W_j^T \mathbf{f}- W_i^T \mathbf{f})}}}.
\end{aligned}
\end{equation}
Recall that  $W_i \mathbf{f}- W_j \mathbf{f} \ge 0$ if $i=\arg\max_j{(W_j \mathbf{f})}$, so $t (W_j^T \mathbf{f}- W_i^T \mathbf{f}) \le 0$ always holds. Then
\begin{equation}
\begin{aligned}
P_i\left({s\mathbf{f}}\right) & \ge \frac{e^{W_i^T \mathbf{f}}}{\sum_{j=1}^{n}{e^{W_j^T \mathbf{f}}}}\\
& = P_i(\mathbf{f}).
\end{aligned}
\end{equation}
The equality holds if $W^T \mathbf{f}=\mathbf{0}$ or $W_i = W_j, \forall i,j \in [1, n]$, which is almost impossible in practice.

\subsection{Proof of Proposition 2}
\label{sec:proof2}
\noindent \textbf{Proposition 2.} (Loss Bound After Normalization)\emph{ Assume that every class has the same number of samples, and all the samples are well-separated, \emph{i.e.} each sample's feature is exactly the same with its corresponding class's weight. If we normalize both the features and every column of the weights to have a norm of $\ell$, the softmax loss will have a lower bound, $\log \left( 1+ \left(n-1 \right)e^{-\frac{n}{n-1}\ell^2}\right)$, where $n$ is the class number. }

\noindent \textbf{Proof:} Assume $\|W_i\|=\ell, \forall i \in [1, n]$ for convenience. Since we have already assumed that all samples are well-separated, we directly use $W_i$ to represent the $i$-th class' feature.

The definition of the softmax loss is,
\begin{equation}
\begin{aligned}
\mathcal{L}_\mathcal{S} = -\frac{1}{n}\sum_{i=1}^n{\log\frac{e^{W_i^T W_i}}{\sum_{j=1}^n{e^{W_i^T W_j}}}}.
\end{aligned}
\end{equation}
This formula is different from Equation (\ref{eq:softmax}) because we assume that every class has the same  sample number. By dividing $e^{W_i^T W_i} = e^{\ell^2}$ from both the numerator and denominator,
\begin{equation}
\begin{aligned}
\mathcal{L}_\mathcal{S} & = -\frac{1}{n}\sum_{i=1}^n{\log\frac{1}{1 + \sum_{j=1,j\neq i}^n{e^{W_i^T W_j - \ell^2}}}}\\
& = \frac{1}{n}\sum_{i=1}^n{\log\left({1 + \sum_{j=1,j\neq i}^n{e^{W_i^T W_j - \ell^2}}}\right)}.
\end{aligned}
\end{equation}
Since $f(x) = e^x$ is a convex function, $\frac{1}{n}\sum_{i=1}^n{e^{x_i}} \ge e^{\frac{1}{n}\sum_{i=1}^{n}{x_i}}$, then we have,
\begin{equation}
\begin{aligned}
\mathcal{L}_\mathcal{S} \ge \frac{1}{n}\sum_{i=1}^n{\log\left({1 + (n-1) e^{\frac{1}{n-1}\sum_{j=1,j\neq i}^n{(W_i^T W_j - \ell^2)}}}\right)}.
\end{aligned}
\end{equation}
The equality holds if and only if all $W_i^T W_j, 1\le i < j \le n$ have the same value, i.e., features from different classes have the same distance. Unfortunately, in $d$-dimension space, there are only $d+1$ unique vertices to ensure that every two vertices have the same distance. All these vertices will form a regular $d$-simplex\cite{rudin1964principles}, e.g., a regular 2-simplex is an equilateral triangle and a regular 3-simplex is a regular tetrahedron. Since the class number is usually much bigger than the dimension of feature in face verification datasets, this equality actually cannot hold in practice. One improvement over this inequality is taking the feature dimension into consideration because we actually have omitted the feature dimension term in this step.

Similar with $f(x) = e^x$, the softplus function $s(x) = \log(1 + C e^x)$ is also a convex function when $C>0$, so that $\frac{1}{n}\sum_{i=1}^n{\log(1 + C e^{x_i})} \ge \log(1 + C e^{\frac{1}{n}\sum_{i=1}^{n}{x_i}})$, then we have

\begin{equation}
\begin{aligned}
\mathcal{L}_\mathcal{S} &\ge \log \left( 1+ \left(n-1 \right)e^{\frac{1}{n(n-1)}\sum_{i=1}^n\sum_{j=1,j\neq i}^n{(W_i^T W_j - \ell^2)}} \right) \\
& = \log \left( 1+ \left(n-1 \right)e^{\left(\frac{1}{n(n-1)}\sum_{i=1}^n\sum_{j=1,j\neq i}^n{W_i^T W_j}\right) - \ell^2} \right).
\end{aligned} 
\end{equation}
This equality holds if and only if $\forall W_i$, the sums of distances to other class' weight $\sum_{j=1,j\neq i}^n{W_i^T W_j}$ are all the same.

Note that
\begin{equation}
\|\sum_{i=1}^n{W_i}\|_2^2 = n\ell^2 + \sum_{i=1}^n\sum_{j=1,j\neq i}^n{W_i^T W_j},
\end{equation}
so
\begin{equation}
\sum_{i=1}^n\sum_{j=1,j\neq i}^n{W_i^T W_j} \ge -n\ell^2.
\end{equation}
The equality holds if and only if $\sum_{i=1}^n{W_i}=\mathbf{0}$. Thus,
\begin{equation}
\begin{aligned}
\mathcal{L}_\mathcal{S} &\ge \log \left( 1+ \left(n-1 \right)e^{-\frac{n\ell^2}{n(n-1)}-\ell^2}\right)\\
& =\log \left( 1+ \left(n-1 \right)e^{-\frac{n}{n-1}\ell^2}\right).
\end{aligned} 
\end{equation}

\subsection{Proof of Proposition 3}
\label{sec:proof3}
\noindent \textbf{Proposition 3.} \emph{ Using an agent for each class instead of a specific sample would cause a distortion of $\frac{1}{n_{\mathcal{C}_i}}\sum_{j \in \mathcal{C}_i}{\left(d(f_0, f_j) - d(f_0, W_i)\right)^2}$, where $W_i$ is the agent of the $i$th-class. The distortion is bounded by $\frac{1}{n_{\mathcal{C}_i}}\sum_{j \in \mathcal{C}_i}{d(f_j, W_i)^2}$.}

\noindent \textbf{Proof:} Since d(x,y) is a metric, through the triangle inequality we have
\begin{equation}
d(f_0,W_i) - d(f_j,W_i) \le d(f_0,f_j) \le d(f_0,W_i) + d(f_j,W_i),
\end{equation}
so
\begin{equation}
- d(f_j,W_i)  \le d(f_0,f_j) - d(f_0,W_i) \le d(f_j,W_i),
\end{equation}
and thus,
\begin{equation}
\left( d(f_0,f_j) - d(f_0,W_i) \right)^2 \le d(f_j,W_i)^2.
\end{equation}
As a result,
\begin{equation}
\begin{aligned}
\frac{1}{n_{\mathcal{C}_i}}\sum_{j \in \mathcal{C}_i}{\left(d(f_0, f_j) - d(f_0, W_i)\right)^2} \le \frac{1}{n_{\mathcal{C}_i}}\sum_{j \in \mathcal{C}_i}{d(f_j, W_i)^2}.
\end{aligned}
\end{equation}

\subsection{Inference of Equation \ref{eq:norm_gradient}}
\label{sec:inference1}
\noindent \textbf{Equation \ref{eq:norm_gradient}}: 
\begin{equation}
\frac{\partial \mathcal{L}}{\partial \mathbf{x}_i} = \frac{\frac{\partial \mathcal{L}}{\partial \tilde{\mathbf{x}}_i} - 
	\tilde{\mathbf{x}}_i\sum_j{\frac{\partial \mathcal{L}}{\partial \tilde{\mathbf{x}}_j}\tilde{\mathbf{x}}_j}}
{\Vert \mathbf{x}  \Vert _2}.
\end{equation}
\textbf{Inference}:
Here we treat $\Vert \mathbf{x}  \Vert _2$ as an independent variable. Note that $\mathbf{\tilde{x}} = \frac{\mathbf{x}}{\Vert \mathbf{x}  \Vert_2}$ and $\Vert \mathbf{x}  \Vert _2 = \sqrt{\sum_j{\mathbf{x}_j^2 +\epsilon}}$ . We have,
\begin{equation}
\begin{aligned}
\frac{\partial \mathcal{L}}{\partial \mathbf{x}_i} & = \frac{\partial \mathcal{L}}{\partial \tilde{\mathbf{x}}_i}\frac{\partial \tilde{\mathbf{x}}_i}{\partial \mathbf{x}_i} + \sum_j{\frac{\partial \mathcal{L}}{\partial \tilde{\mathbf{x}}_j}\frac{\partial \tilde{\mathbf{x}}_j}{\partial \Vert \mathbf{x}  \Vert _2}\frac{\partial \Vert \mathbf{x}  \Vert _2}{\partial \mathbf{x}_i}}\\
& = \frac{\partial \mathcal{L}}{\partial \mathbf{\tilde{x}}}\frac{1}{\Vert \mathbf{x}  \Vert_2} + \sum_j{\frac{\partial \mathcal{L}}{\partial \tilde{\mathbf{x}}_j}\frac{-\tilde{\mathbf{x}}_j}{\Vert \mathbf{x}  \Vert_2^2}\cdot \frac{1}{2}\frac{1}{\Vert \mathbf{x}  \Vert_2}}\cdot 2\mathbf{x}_i\\
& = \frac{\partial \mathcal{L}}{\partial \mathbf{\tilde{x}}}\frac{1}{\Vert \mathbf{x}  \Vert_2} - \frac{\mathbf{x}_i}{\Vert \mathbf{x}  \Vert_2}}\sum_j{\frac{\partial \mathcal{L}}{\partial \tilde{\mathbf{x}}_j}\frac{\tilde{\mathbf{x}}_j}{\Vert \mathbf{x}  \Vert_2^2}\\
& = 
\frac{\frac{\partial \mathcal{L}}{\partial \tilde{\mathbf{x}}_i} - 
	\tilde{\mathbf{x}}_i\sum_j{\frac{\partial \mathcal{L}}{\partial \tilde{\mathbf{x}}_j}\tilde{\mathbf{x}}_j}}
{\Vert \mathbf{x}  \Vert _2}.
\end{aligned}
\end{equation}

\subsection{Proof of $\langle \mathbf{x}, \frac{\partial \mathcal{L}}{\partial \mathbf{x}} \rangle = 0$}
\label{sec:proof4}
\noindent \textbf{Proof}: The vectorized version of Equation \ref{eq:norm_gradient} is
\begin{equation}
\frac{\partial \mathcal{L}}{\partial \mathbf{x}} = \frac{\frac{\partial \mathcal{L}}{\partial \tilde{\mathbf{x}}} - 
	\tilde{\mathbf{x}}\langle \frac{\partial \mathcal{L}}{\partial \tilde{\mathbf{x}}}, \tilde{\mathbf{x}}\rangle} 
{\Vert \mathbf{x}  \Vert _2}.
\end{equation}
So,
\begin{equation}
\begin{aligned}
\langle \mathbf{x}, \frac{\partial \mathcal{L}}{\partial \mathbf{x}} \rangle & = \frac{\langle \mathbf{x}, \frac{\partial \mathcal{L}}{\partial \tilde{\mathbf{x}}}\rangle - \langle \mathbf{x}, 
	\tilde{\mathbf{x}}\rangle \langle \frac{\partial \mathcal{L}}{\partial \tilde{\mathbf{x}}}, \tilde{\mathbf{x}}\rangle} 
{\Vert \mathbf{x}  \Vert _2} \\
& =  \frac{\langle \mathbf{x}, \frac{\partial \mathcal{L}}{\partial \tilde{\mathbf{x}}}\rangle - \frac{\langle \mathbf{x}, 
	\mathbf{x}\rangle \langle \frac{\partial \mathcal{L}}{\partial \tilde{\mathbf{x}}}, \mathbf{x}\rangle}{\Vert \mathbf{x}  \Vert _2^2}} 
{\Vert \mathbf{x}  \Vert _2}\\
& = \frac{\langle \mathbf{x}, \frac{\partial \mathcal{L}}{\partial \tilde{\mathbf{x}}}\rangle - \langle \tilde{\mathbf{x}}, 
	\tilde{\mathbf{x}}\rangle \langle \frac{\partial \mathcal{L}}{\partial \tilde{\mathbf{x}}}, \mathbf{x}\rangle} 
{\Vert \mathbf{x}  \Vert _2}\\
& = \frac{\langle \mathbf{x}, \frac{\partial \mathcal{L}}{\partial \tilde{\mathbf{x}}}\rangle - \langle \frac{\partial \mathcal{L}}{\partial \tilde{\mathbf{x}}}, \mathbf{x}\rangle} 
{\Vert \mathbf{x}  \Vert _2}\\
& = 0.
\end{aligned}
\end{equation}
\end{document}